%% file: example_paper.tex
\pdfoutput=1

\documentclass{article}

\input{math_commands.tex}

\usepackage{xcolor}
\usepackage{colortbl}
\usepackage{hyperref}
\usepackage{url}
\usepackage{amsmath}
\usepackage{amsthm}
\usepackage{amssymb}
\usepackage{dsfont}
\usepackage[capitalize,nameinlink]{cleveref}[0.19]
\usepackage{enumerate}
\usepackage[shortlabels]{enumitem}
\allowdisplaybreaks
\usepackage{mdframed}
\usepackage[most]{tcolorbox}   

\usepackage{graphicx}
\usepackage{svg}
\usepackage{microtype}
\usepackage{booktabs} 

\usepackage{caption}
\usepackage{subcaption}

\usepackage{booktabs, multicol, multirow, diagbox}

\newtheorem{theorem}{Theorem}
\newtheorem{lemma}[theorem]{Lemma}

\newtheorem{corollary}[theorem]{Corollary}
\newtheorem{definition}{Definition}

\newtheorem{assumption}{Assumption}

\newtheorem{remark}{Remark}

\crefname{definition}{Definition}{Definitions}
\crefname{assumption}{Assumption}{Assumptions}
\crefname{theorem}{Theorem}{Theorems}
\crefname{remark}{Remark}{Remarks}
\crefname{lemma}{Lemma}{Lemmas}
\crefname{corollary}{Corollary}{Corollaries}
\crefname{proposition}{Proposition}{Propositions}
\crefname{section}{Section}{Sections}
\crefname{subsection}{Subsection}{Subsections}
\crefname{example}{Example}{Examples}
\crefname{table}{Table}{Tables}
\crefname{problem}{Problem}{Problems}
\crefname{algorithm}{Algorithm}{Algorithms}
\crefname{figure}{Figure}{Figures}
\crefname{property}{Property}{Properties}

\crefformat{equation}{#2Equation~(#1)#3}
\crefformat{inequality}{#2Inequality~{}(#1){}#3}
\crefrangeformat{inequality}{#3Inequality~(#1)#4--#5(#2)#6}
\Crefformat{section}{#2Section~#1#3}
\Crefformat{page}{#2Page~#1#3}

\newcommand{\bw}{\mathbf{w}}
\newcommand{\bW}{\mathbf{W}}
\newcommand{\bZ}{\mathbf{Z}}

\newcommand{\bp}{\mathbf{P}}

\newcommand{\bM}{\mathbf{M}}

\newcommand{\xcal}{\mathcal{X}}
\newcommand{\ycal}{\mathcal{Y}}
\newcommand{\scal}{\mathcal{S}}
\newcommand{\dcal}{\mathcal{D}}

\newcommand{\zcal}{\mathcal{Z}}
\newcommand{\ebb}{\mathbb{E}}
\newcommand{\rbb}{\mathbb{R}}

\newcommand*{\tran}{^{\mkern-1.5mu\mathsf{T}}}
\newcommand{\ind}{\mathds{1}}
\newcommand{\unaryminus}{\scalebox{0.75}[1.0]{\( - \)}}
\newcommand{\unaryplus}{\scalebox{0.75}[1.0]{\( + \)}}

\usepackage[normalem]{ulem}

\makeatletter
\newcommand{\biggg}{\bBigg@{3}}
\newcommand{\vast}{\bBigg@{4}}
\makeatother

\usepackage[accepted]{icml2022}

\icmltitlerunning{Topology-aware Generalization of Decentralized SGD}

\begin{document}

\twocolumn[
\icmltitle{Topology-aware Generalization of Decentralized SGD}



\icmlsetsymbol{equal}{*}

\begin{icmlauthorlist}
\icmlauthor{Tongtian Zhu}{zju,sias,jde}
\icmlauthor{Fengxiang He}{jde}
\icmlauthor{Lan Zhang}{ustc,hefei}
\icmlauthor{Zhengyang Niu}{whu,jde}
\icmlauthor{Mingli Song}{zjuc}
\icmlauthor{Dacheng Tao}{jde}
\end{icmlauthorlist}

\icmlaffiliation{zju}{College of Computer Science and Technology, Zhejiang University}
\icmlaffiliation{sias}{Shanghai Institute for Advanced Study of Zhejiang University}
\icmlaffiliation{jde}{JD Explore Academy, JD.com Inc.}
\icmlaffiliation{ustc}{School of Computer Science and Technology, University of Science and Technology of China}
\icmlaffiliation{hefei}{Institute of Artificial Intelligence, Hefei Comprehensive National Science Center}
\icmlaffiliation{whu}{School of Computer Science, Wuhan University}
\icmlaffiliation{zjuc}{Zhejiang University City College} 


\icmlcorrespondingauthor{Fengxiang He}{fengxiang.f.he@gmail.com}

\icmlkeywords{Generalization Theory, Decentralized Learning}

\vskip 0.3in
]



\printAffiliationsAndNotice{}  

\input{section/0-abstract}
\input{section/1-introduction}
\input{section/2-related_work} 
\input{section/3-preliminaries}

\input{section/4-theoretical_results}

\input{section/5-empirical_results}
\input{section/6-discussion}
\input{section/7-conclusion}
\input{section/acknowledgement}
\input{section/appendix}

\end{document}

%% file: math_commands.tex
\usepackage{amsmath,amsfonts,bm}

\def\eqref#1{equation~\ref{#1}}

\def\1{\bm{1}}

\DeclareMathAlphabet{\mathsfit}{\encodingdefault}{\sfdefault}{m}{sl}
\SetMathAlphabet{\mathsfit}{bold}{\encodingdefault}{\sfdefault}{bx}{n}

\DeclareMathOperator{\Tr}{Tr}

%% file: section/0-abstract.tex
\begin{abstract}
    This paper studies the algorithmic stability and generalizability of decentralized stochastic gradient descent (D-SGD). We prove that the consensus model learned by D-SGD is $\mathcal{O}{(N^{-1}\unaryplus m^{-1} \unaryplus\lambda^2)}$-stable in expectation in the non-convex non-smooth setting, where $N$ is the total sample size, $m$ is the worker number, and $1\unaryminus\lambda$ is the spectral gap that measures the connectivity of the communication topology. These results then deliver an $\mathcal{O}{(N^{-(1\unaryplus\alpha)/2}\unaryplus m^{-(1+\alpha)/2}\unaryplus\lambda^{1+\alpha} \unaryplus \phi_\scal)}$ in-average generalization bound, which is non-vacuous even when $\lambda$ is closed to $1$, in contrast to vacuous as suggested by existing literature on the projected version of D-SGD. Our theory indicates that the generalizability of D-SGD is positively correlated with the spectral gap, and can explain why consensus control in initial training phase can ensure better generalization. Experiments of VGG-11 and ResNet-18 on CIFAR-10, CIFAR-100 and Tiny-ImageNet justify our theory. To our best knowledge, this is the first work on the topology-aware generalization of vanilla D-SGD. Code is available at \url{https://github.com/Raiden-Zhu/Generalization-of-DSGD}.
\end{abstract}

%% file: section/1-introduction.tex
\section{Introduction}
Decentralized stochastic gradient descent (D-SGD) 
facilitates simultaneous model training on a massive number of workers without a central server \citep{lopes2008diffusion, nedic2009distributed}. In D-SGD, every worker only communicates with the directly connected neighbors through “gossip communication'' \citep{xiao2004fast,NIPS2017_f7552665, pmlr-v119-koloskova20a}. The communication intensity is controlled by the communication topology. This decentralized nature eliminates the requirement for an expensive central server dedicated to heavy communication. Surprisingly, existing theoretical results demonstrate that the massive models on the edge converge to a unique steady model, the consensus model, even without the control of a central server \citep{lu2011gossip, shi2015extra, NIPS2017_f7552665}. Compared with the centralized synchronized SGD (C-SGD) \citep{dean2012large,li2014communication}, D-SGD can achieve the same asymptotic linear speedup in convergence rate \citep{NIPS2017_f7552665}. 
In this way, D-SGD provides a promising distributed machine learning paradigm with improved privacy \citep{nedic2020distributed}, scalability \citep{NIPS2017_f7552665,9464278}, and communication efficiency \citep{ying2021bluefog}.

To date, the theoretical research on D-SGD has mainly focused on its convergence \citep{nedic2009distributed, NIPS2017_f7552665, pmlr-v119-koloskova20a,alghunaim2021unified}, while the understanding on the generalizability \citep{mohri2018foundations,he2020recent} of D-SGD is still premature. A large amount of empirical evidence have shown that D-SGD generalizes well on well-connected topologies \citep{assran2019stochastic,ying2021exponential}.
Meanwhile, empirical results by \citet{pmlr-v97-assran19a}, \citet{ pmlr-v139-kong21a} and \citet{ying2021exponential} demonstrate that for ring topologies, the validation accuracy of the consensus model learned by D-SGD decreases as the number of workers increases. Thus, a question is raised:
\begin{tcolorbox}[notitle, sharp corners, colframe=darkgrey, colback=white, 
       boxrule=1pt, boxsep=0.5pt, enhanced, 
       shadow={3pt}{-3pt}{0pt}{opacity=1,mygrey}, title={Our question},]
\emph{{How does the communication topology of D-SGD impact its generalizability?}}
\end{tcolorbox}
This paper answers this question. We prove a topology-aware generalization error bound for the consensus model learned by D-SGD, 
which characterizes the impact of the communication topology on the generalizability of D-SGD.
Our contributions are summarized as follows:
\begin{itemize}[leftmargin=*]
    \item \textbf{Stability and generalization bounds of D-SGD.} This work proves the algorithmic stability \citep{bousquet2002stability} and generalization bounds of vanilla D-SGD in the non-convex non-smooth setting. 
    In \cref{sec:theoretical results}, we present an $\mathcal{O}{(N^{-1}\unaryplus m^{-1}\unaryplus\lambda^2)}$ distributed on-average stability (see \cref{th:stab-fix-eta}), where $1-\lambda$ denotes the spectral gap of the network, a measure of the connectivity of the communication topology $\mathcal{G}$. 
   These results would suffice to derive a $\mathcal{O}{(N^{-(1\unaryplus\alpha)/2}\unaryplus m^{-(1+\alpha)/2}\unaryplus\lambda^{1+\alpha} \unaryplus \phi_\scal)}$ generalization bound in expectation of D-SGD (see \cref{thm:gen-expected}). Our error bounds are non-vacuous, even when the worker\footnote{Throughout this work, we use the term \textit{worker} to represent the local model.} number is sufficiently large, or the communication graph is sufficiently sparse. The theory can be directly applied 
   to explain why consensus distance control in the initial phase of training can ensure better generalization.

    \item  \textbf{Communication topology and generalization of D-SGD.} Our theory shows that the generalizability of D-SGD has a positive relationship with the spectral gap $1-\lambda$ of the communication topology $\mathcal{G}$. 
    Besides, we prove that the generalizability of D-SGD decreases when the worker number increases for the ring, grid, and exponential graphs. We conduct comprehensive experiments of VGG-11 \citep{simonyan2014very} and ResNet-18 \citep{he2016identity} on CIFAR-10,  CIFAR-100 \citep{krizhevsky2009learning} and Tiny-ImageNet \citep{le2015tiny} to verify our theory.
\end{itemize}

To our best knowledge, this work offers the first investigation into the topology-aware generalizability of vanilla D-SGD. The closest work in the existing literature is by \citet{Sun_Li_Wang_2021}, which derives $\mathcal{O}({N}^{-1}\unaryplus (1\unaryminus\lambda)^{-1})$ generalization bounds for projected D-SGD based on uniform stability \citep{bousquet2002stability}. They show that the decentralized nature hurts the stability, and thus undermines generalizability. 
Compared with the results by \citet{Sun_Li_Wang_2021}, our work makes two contributions: (1) we analyze the vanilla D-SGD, which is capable of solving optimization problems on unbounded domains, rather than the projected D-SGD; and (2) our stability and generalization bounds are non-vacuous, even in the cases where the spectral gap $1\unaryminus\lambda$ is sufficiently close to $0$, which characterizes the cases where the worker number is sufficiently large or the communication graph is sufficiently sparse. 

%% file: section/2-related_work.tex
\section{Related Work}

       The earliest work of classical decentralized optimization can  be  traced  back  to \citet{tsitsiklis1984problems}, \citet{afe727cd65e44d8ebb1d1e1ad602b087} and \citet{4749425}. 
       D-SGD 
       has been extended to various settings in deep learning, including time-varying topologies \citep{lu2020decentralized,pmlr-v119-koloskova20a}, asynchronous settings \citep{lian2018asynchronous,xu2021dp,nadiradze2021asynchronous}, directed topologies \citep{assran2019stochastic,taheri2020quantized}, and data-heterogeneous scenarios \citep{tang2018d,vogels2021relaysum}. It has been proved that the convergence of D-SGD heavily relies on the communication topology \cite{hambrick1996influence,bianchi2012convergence,NIPS2017_f7552665, nedic2018network, pmlr-v97-assran19a,wang2019matcha,guo2020communication, NEURIPS2022beyond}, especially in the scenarios where the local data is heterogeneous across workers \citep{9139399,pmlr-v119-koloskova20a, bellet2021d, dai2022dispfl, bars2022yes}. However, the impact of the communication topology on the generalizability of D-SGD is still in its infancy.
      
      Recently, inspiring work by \citet{zhang2021loss} gives insights to how gossip communication in D-SGD promotes generalization in large batch settings. 
      They prove that a self-adjusting noise exists in D-SGD, which may help D-SGD find flatter minima with better generalization.
    Another work by \citet{richards2020graph} presents a generalization bound of the Adaptation-Then-Combination (ATC) version of D-SGD through algorithmic stability and Rademacher complexity \citep{mohri2018foundations} in both smooth and non-smooth settings. However, their generalization bounds are invariant  to the communication topology, which contradicts the experimental results (see \cref{fig:topology-comparistion}). 
    In contrast, our generalization bounds are topology-aware and characterize the effects of decentralization on generalization.


%

%% file: section/3-preliminaries.tex
\section{Preliminaries}

\textbf{Supervised learning.} 
Supposed $\xcal \subseteq \rbb^{d_x}$ and $\ycal \subseteq \rbb$ are the input and output spaces, respectively. We denote the training set as $\scal=\left\{z_{1}, \ldots, z_{N}\right\}$, where $z_{\zeta}=\left(x_{\zeta}, y_{\zeta}\right), \zeta=1,\dots, N $ are sampled independent and identically distributed (i.i.d.) from an unknown data distribution $\dcal$ defined on $\mathcal{Z}=\mathcal{X} \times \mathcal{Y}$. 

The goal of supervised learning is to learn a predictor (hypothesis) $g(\cdot ;\bw)$, parameterized by $\bw=\bw(z_{1}, z_{2}, \ldots, z_{N}) \in \rbb^d$, to approximate the mapping between the input variable $x\in \mathcal{X}$ and the output variable $y\in \mathcal{Y}$, based on the training set $\scal$. Let $c: \mathcal{Y} \times \mathcal{Y} \mapsto \mathbb{R}^{+}$ be a loss function that evaluates the prediction performance of the hypothesis $g$. The loss of a hypothesis $g$ with respect to (w.r.t.) the example $z_{\zeta}=(x_{\zeta}, y_{\zeta})$ is denoted by $f(\bw; z_{\zeta})=c(g(x_{\zeta};\bw), y_{\zeta})$, in order to measure the effectiveness of the learned model. Then, the empirical and population risks of $\bw$ are defined as follows:
\begin{equation*}\label{eq:risk}
    F_\scal(\bw)=\frac{1}{N}\sum_{\zeta=1}^{N}f(\bw;z_\zeta),~~  F(\bw)=\ebb_{z\sim D}[f(\bw;z)].
\end{equation*}

\begin{figure*}[!t]
\vskip 0.2in
\begin{center}
\centerline{\includegraphics[width=0.95\textwidth]{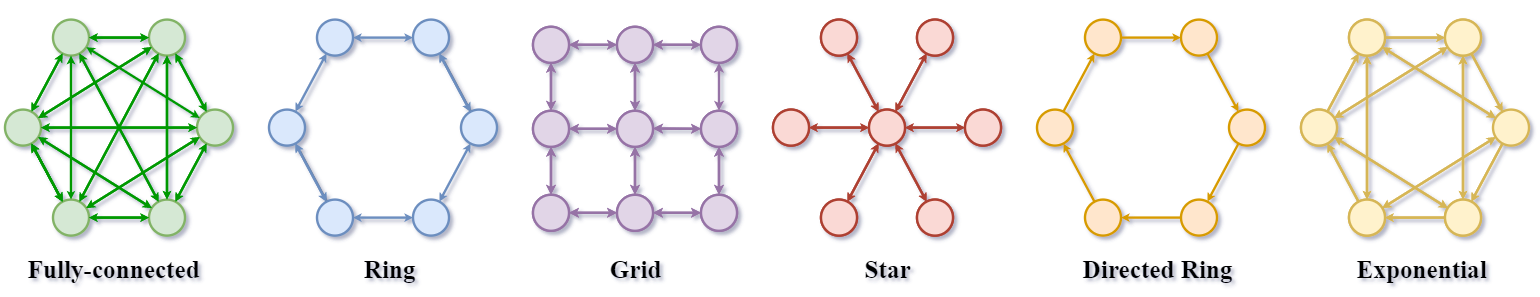}}
\caption{Illustration of some commonly-used communication topologies.} 
\label{fig:topologies}
\end{center}
\vskip -0.2in
\end{figure*}
\textbf{Distributed learning.}\label{def:dis-setting}
Distributed learning jointly trains a learning model $\bw$ on multiple workers 
\citep{shamir2014distributed}. In this framework, the $k$-th worker $(k=1,\dots, m)$ can access $n_{k}$ independent and identically distributed (i.i.d.) training examples $\scal_k=\{z_{k,1}, \ldots, z_{k,n_{k}}\}$, drawn from the data distribution $\mathcal{D}$. If set $n_{k}=n$, the total sample size will be $N=n m$. In this case, the global empirical risk of $\bw$ is
\begin{equation*}
\hat{F}(\bw) = \frac{1}{m}\sum_{k=1}^{m}F_{\scal_k}(\bw),
\end{equation*}
where $F_{\scal_k}(\bw)=\frac{1}{n}\sum_{\zeta=1}^{n}f({\bw};z_{k, \zeta})$
denotes the local empirical risk on the $k$-th worker and $z_{k, \zeta}\in\scal_{k}$ $(\zeta =1,\dots, n)$ is the local sample set.

\textbf{Decentralized Stochastic Gradient  Descent (D-SGD). }\label{def:dec-sgd}
The goal of D-SGD is to learn a consensus model $\overline{\bw}=\frac{1}{m}\sum_{k=1}^m\bw_{k}$, on $m$ workers, where $\bw_k$ denotes the local model on the $k$-th worker. 
For any $k$, let $\bw_k^{(t)}\in\rbb^d$ be the $d$-dimensional local model on the $k$-th worker in the $t$-th iteration, while $\bw_{k}^{(1)}=0$ is the initial point. 
We denote $\bp$ as a doubly stochastic gossip matrix that characterizes the underlying topology $\mathcal{G}$ (see \cref{def:dou-matrix} and \cref{fig:topologies}). The intensity of gossip communications is measured by the spectral gap \citep{seneta2006non} of $\bp$ (i.e., $1-\max \left\{\left|\lambda_{2}\right|,\left|\lambda_{n}\right|\right\}$, where $\lambda_i\ (i=2,\dots,m)$ denotes the $i$-th largest eigenvalue of $\bp$ (see \cref{def:spectral-gap}). The vanilla Adapt-While-Communicate  (AWC) version of D-SGD without projecting operations updates the model on the $k$-th worker by

\begin{equation}\label{eq:dec-sgd-entry}
    \bw_{k}^{(t+1)}=\overbrace{\sum_{l=1}^{m}\bp_{k,l} \bw_l^{(t)}}^{\text Communication}-\overbrace{\eta_t\nabla f(\bw_k^{(t)};z_{k,\zeta_t}^{(t)})}^{\text Computation},
\end{equation}
where $\{\eta_t\}$ is a sequence of positive learning rates, and $\nabla f(\bw_k^{(t)};z_{k,\zeta_t}^{(t)})$ is the gradient of $f$ w.r.t. the first argument on the $k$-th worker, and $\zeta_t$ is i.i.d. variable drawn from the uniform distribution over $\{1,\ldots,n\}$ at the $t$-th iteration \citep{NIPS2017_f7552665}.
In this paper, matrix $\bW= [\bw_{1}, \cdots, \bw_{m}]^T \in \rbb^{m\times d}$ stands for all local models across the network, while matrix $\nabla f(\bW;\bZ) = [\nabla f(\bw_{1};z_{1}), \cdots, \nabla f(\bw_{m};z_{m})]^T \in \rbb^{m\times d}$ stacks all local gradients w.r.t. the first argument.
In this way, the matrix form of \cref{eq:dec-sgd-entry} is as follows:
\begin{equation*}\label{eq:dec-sgd-matrix}
    \bW^{(t+1)}=\bp\bW^{(t)}-\eta_t\nabla f(\bW^{(t)};\bZ^{(t)}_{\zeta_t}).
\end{equation*}

%% file: section/4-theoretical_results.tex
\section{Topology-aware Generalization Bounds of D-SGD}\label{sec:theoretical results}

This section proves stability and generalization bounds for 
D-SGD. 
We start with the definition of a new parameter-level stability for distributed settings. 
Then, the stability of D-SGD under a non-smooth condition is obtained (see \cref{thm:on-average-holder} and \cref{th:stab-fix-eta}).
This implies a connection between stability and generalization in expectation (see \cref{prop:gen-stab}), which suffices to prove the expected generalization bound of D-SGD, of order $\mathcal{O}{(N^{-(1\unaryplus\alpha)/2}\unaryplus m^{-(1+\alpha)/2}\unaryplus\lambda^{1+\alpha} \unaryplus \phi_\scal)}$.

\subsection{Algorithmic Stability of D-SGD}

Understanding generalization using algorithmic stability can be traced back to \citet{bousquet2002stability} and \citet{shalev2010learnability}, and has been applied to stochastic gradient methods \citep{pmlr-v48-hardt16,pmlr-v119-lei20c}. For more details, please see \cref{sec:add-related-work}.

We define a new algorithmic stability of distributed optimization algorithms below, which better characterizes the on-average sensitivity of models across multiple workers.

\begin{definition}[Distributed On-average Stability\label{def:dis-aver-stab}]
Let $\scal_k=\{z_{k,1}, \ldots, z_{k,n}\}$ denote the i.i.d. local samples on $k$-th worker drawn from the distribution $\dcal$. $\scal=\cup_{k=1}^m \scal_k=\{z_{1}, \ldots, z_{N}\}$ then denotes the whole training set. $\scal^{(i)}=\cup_{k=1}^m \scal_k^{(i)}=\{z_{1}, \ldots,\tilde{z}_{i},\ldots, z_{N}\}$ is formed by replacing the $i$-th element of $\scal$ with a sample $\tilde{z}_{i}$ drawn from the distribution $\dcal$, where $\scal_k^{(i)}$ denotes the new local training samples on $k$-th worker\footnote{Note that $\scal_k$ and $\scal_k^{(i)}$ can be exactly the same with probability $1-\frac{1}{m}$, since the only one data point replaced can be located in any of the $m$ local data sets with equal probability.}. We denote $\bw_k$ and $\widetilde{\bw}_k$ as the weight vectors on the $k$-th worker produced by the stochastic algorithm $A$ based on $\scal$ and $\scal^{(i)}$, respectively.
$A$ is $\ell_2$ distributed on-average $\epsilon$-stable for all training data sets $\scal$ and $\scal^{(i)}$ if
\begin{equation*}
    \frac{1}{mN}\sum_{i=1}^{N}\sum_{k=1}^{m}\ebb_{\scal,\scal^{(i)},A}\big[\|\bw_k-\widetilde{\bw}_k\|_2^2\big]\leq\epsilon^2,
\end{equation*}
where $\ebb_A[\cdot]$ stands for the expectation w.r.t. the randomness of the algorithm $A$ (see more details in \cref{sec:add-back-ground}).
\end{definition}

We then prove that D-SGD is distributed on-average stable.

\begin{theorem}
\label{thm:on-average-holder}
  Let $\scal_k$ and $\scal^{(i)}_k$ $\ (k=1,\dots, m)$ be constructed in \cref{def:dis-aver-stab}. 
  Let $\bw^{(t)}_k$ and $\widetilde{\bw}^{(t)}_k$ be the $t$-th iteration on the $k$-th worker produced by \cref{eq:dec-sgd-entry} based on $\scal_k$ and $\scal_k^{(i)}$ $ (k=1,\dots,m)$ respectively, and $\{\eta_t\}$ be a non-increasing sequence of positive learning rates. 
  We assume that for all $z\in\zcal$, the function $\bw\mapsto f(\bw;z)$ is non-negative with its gradient $\nabla f(\bw;z)$ being $(\alpha,L)$-H\"older continuous (see \cref{def:holder}).
  We further assume that the weight differences at the $t$-th iteration are multivariate normally distributed: $\bw_{k}^{(t)}-\widetilde{\bw}_k^{(t)}\stackrel{ i.i.d.}{\sim} \mathcal{N}(\mu_{t,k}, \sigma_{t,k}^2 I_{d})$ for all $k$ where $d$ denotes the dimension of weights, with \textbf{unknown} parameters $\mu_{t,k}$ and $\sigma_{t,k}$ satisfying some technical conditions (see \cref{ass:gaussian-diff-w}), and the worker number $m\geq \frac{1}{d\mu_0^2}$\footnote{$d\mu_0^2$ is the lower bound of $ \|\mu_{t,k}\|_2^2\ (k=1\dots m)$. $m\geq \frac{1}{d\mu_0^2}$ can be easily satisfied in training overparameterized models in a decentralized manner, since both $m$ and $d$ are large in these cases.}. Then we have the following:
\begin{align*}
        \frac{1}{mN}&\sum_{i=1}^{N}\sum_{k=1}^{m}\ebb_{\scal,\scal^{(i)},A}\big[\big\|\bw_{k}^{(t+1)}-\widetilde{\bw}_{k}^{(t+1)}\big\|_{2}^{2}\big]\nonumber\\
        \leq &\sum_{\tau=0}^{t} {C}^{t-\tau}\{\underbrace{\mathcal{O}{\big( (1-\frac{1}{m})\boldsymbol{\lambda^2} + \frac{1}{m}\big)}}_{\text {Error from decentralization}}\nonumber\\ &+\underbrace{\mathcal{O}{(\frac{{\eta_\tau}^2}{N}}\cdot\frac{1}{m}\sum_{k=1}^{m}\ebb_{\scal,A}\big[F_{\scal}^{\frac{2\alpha}{1+\alpha}}(\bw_k^{(\tau)})\big])}_{\mathcal{O}{(\frac{1}{N})}\text { $\cdot$ Averaged empirical risk}}
        \},
\end{align*}
where $C = 2\eta_0 L (1-\frac{1}{N})$ and $F_{\scal}(\bw_k^{(\tau)})$ is the local empirical risk of the $k$-th worker at iteration $\tau$. 
\end{theorem}
 \cref{thm:on-average-holder} suggests that the distributed on-average stability of decentralized SGD is positively related to the spectral gap of the given topology and negatively related to the accumulation of the averaged empirical risk. See detailed proof in \cref{sec:proof-holder-stab}.

We can obtain a simplified result with fixed learning rates.

\begin{corollary}[Stability in Expectation with $\eta_t\equiv\eta$ \label{th:stab-fix-eta}]
    Suppose all the assumptions of \cref{thm:on-average-holder} hold. With a fixed learning rate $\eta_t\equiv\eta \leq \frac{1}{2L} (1-\frac{2}{m})$, the distributed on-average stability of D-SGD can be bounded as
\begin{align*}
        \frac{1}{mN}&\sum_{i=1}^{N}\sum_{k=1}^{m}\ebb_{\scal,\scal^{(i)},A}\big[\big\|\bw_{k}^{(t+1)}-\widetilde{\bw}_{k}^{(t+1)}\big\|_{2}^{2}\big]\nonumber\\
        \leq & \frac{1}{1\unaryminus 2\eta L (1\unaryminus\frac{1}{N})}\{\mathcal{O}{(\frac{{\epsilon_{\scal} \eta}^2}{N})}
         +\underbrace{\mathcal{O}{\big( (1-\frac{1}{m})\boldsymbol{\lambda^2} + \frac{1}{m}\big)}}_{\text {Error from decentralization}}
        \},
\end{align*}
 where $\epsilon_{\scal}$ denotes the upper bound of averaged empirical risk $ \frac{1}{m}\sum_{k=1}^{m}\ebb_{\scal,A}\big[F_{\scal}^{\frac{2\alpha}{1+\alpha}}(\bw_{k}^{(t)})\big]$. 
\end{corollary}
\cref{th:stab-fix-eta} shows that the distributed on-averge stability of D-SGD is of the order $\mathcal{O}{(N^{-1}\unaryplus m^{-1}\unaryplus\lambda^2)}$. We defer the proof to \cref{sec:proof-holder-stab}.

\begin{figure}[t!]
\vskip 0.2in
\begin{center}
\centerline{\includegraphics[width=\columnwidth]{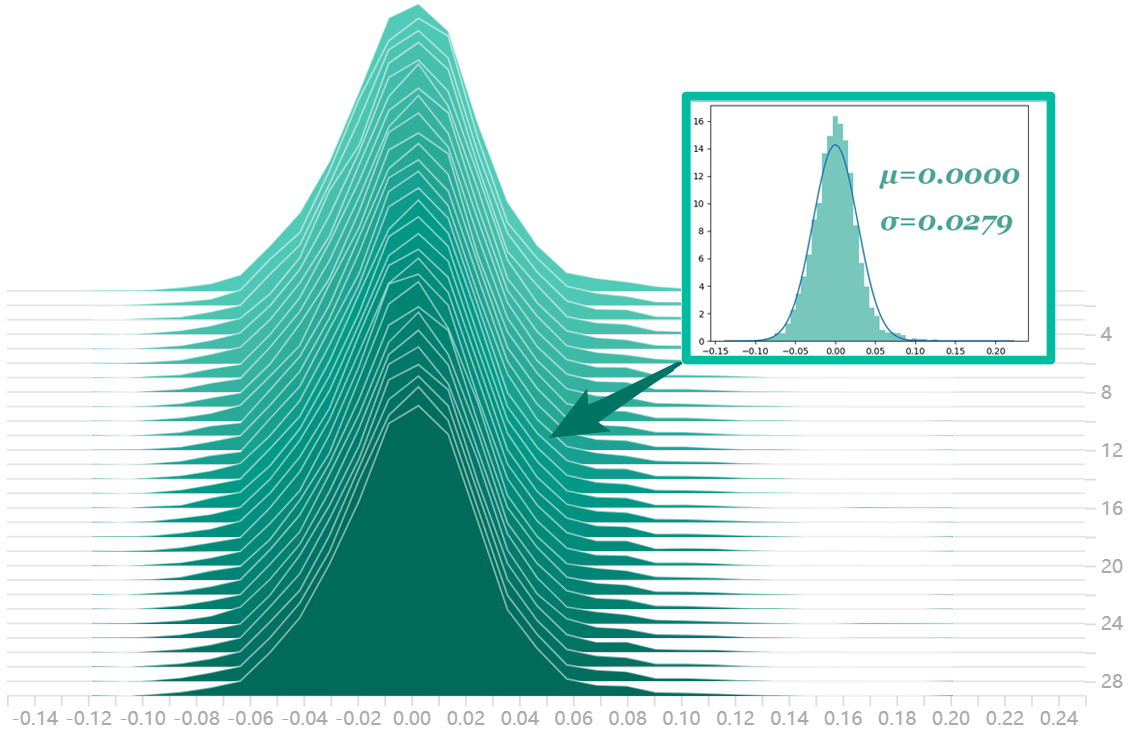}}
\caption{Histograms of the weight differences of the last layers of the ResNet-18 models (1024 dimensions $\times$ 10 classes=10240 parameters) trained by AWC D-SGD on $\scal$ and $\scal^{(i)}$ that differ by only one data point. Thirty ResNet-18 models are trained on data sampled from the MNIST dataset \citep{lecun1998gradient}, each model is trained with 16 workers for 3000 iterations.} 
\label{fig:The distribution on weight differences}
\end{center}
\vskip -0.2in
\end{figure}

\paragraph{Comparison with existing results.} Compared with \citet{Sun_Li_Wang_2021}, we relax the restrictive bounded gradient and the smoothness assumptions. Instead, a much weaker H\"older condition (see \cref{def:holder}) is adopted.
In addition, we make a mild assumption that the weight difference $\big(\bw_{k}^{(t)}-\widetilde{\bw}_k^{(t)}\big)$ is multivariate normally distributed (see \cref{ass:gaussian-diff-w}), which stems from our empirical observations: \cref{fig:The distribution on weight differences} illustrates that the distribution of the weight differences in  ResNet-18 models trained by D-SGD is close to a centered Gaussian.  Intuitively, the assumption is based upon the fact that the weights of the consensus model are very insensitive to the change of a single data point.

We also compare the order of the derived bound with the existing literature.
\citet{pmlr-v48-hardt16} proves that SGD is $\mathcal{O}{(\sum_{\tau=1}^{t-1} \eta_{\tau} /N)}$-stable in convex and smooth settings, which corresponds to the $\mathcal{O}{(\frac{1}{N})}$ term in \cref{th:stab-fix-eta}.
Under H\"older continuous condition, \citet{pmlr-v119-lei20c} proposes a parameter-level stability bound of SGD of the order $\mathcal{O}{(\frac{{\epsilon_{\scal} \eta}^2}{N} \unaryplus \eta^{\frac{2}{1\unaryminus\alpha}})}$. In contrast, \cref{th:stab-fix-eta} shows that D-SGD suffers from additional terms $\mathcal{O}{\big( (1\unaryminus\frac{1}{m})\lambda^2 \unaryplus \frac{1}{m}\big)}$, where the first term $\mathcal{O}{\big( (1\unaryminus\frac{1}{m})\lambda^2})$ can characterize the degree of disconnection of the underlying communication topology.
Close work by \citet{Sun_Li_Wang_2021} proved that the stability of the projected variant of D-SGD is bounded by $\mathcal{O}\left(\frac{\eta t B^2}{N}\unaryplus\frac{\eta t B^2}{1\unaryminus\lambda}\right)$ in the convex smooth setting, where $B$ is the upper bound of the gradient norm. The term $\mathcal{O}{(\frac{\eta t B^2}{1\unaryminus\lambda})}$ brought by decentralization is of the order $\mathcal{O}{(m^2 B^2)}$ for ring topologies and $\mathcal{O}{(m B^2)}$ for grids, respectively. Our error bound  in \cref{th:stab-fix-eta} is tighter than their results, since
\begin{multline*}
\frac{1}{1\unaryminus 2\eta L (1\unaryminus\frac{1}{N})}\cdot\mathcal{O}{\big( (1-\frac{1}{m})\lambda^2 + \frac{1}{m}\big)}\\ \leq \mathcal{O}{\big(m\big)} \ll \mathcal{O}{(m B^2)} \leq \mathcal{O}{(m^2 B^2)}.
\end{multline*}

\subsection{Generalization bounds of D-SGD}

The following lemma bridges the gap between generalization and the newly proposed distributed on-average stability. 
\begin{lemma}[Generalization via Distributed On-average Stability] \label{prop:gen-stab}
    Let $\scal_k$ and $\scal^{(i)}_k$ be constructed in \cref{def:dis-aver-stab}\footnote{We appreciate Xiaolin Hu's comment regarding $\scal_k$ and $\scal^{(i)}_k$.}. If $\forall z$ the pre-specified function $f(\bw;z)$ is non-negative, with its gradient $\nabla f(\bw;z)$ being $(\alpha,L)$-H\"older continuous,\footnote{We appreciate Batiste Le Bars for pointing out an issue about this assumption. The issue has been addressed.} and $\forall t$ $\ebb_{\scal,\scal^{(i)},A}\|\frac{1}{m}\sum_{k=1}^m\bw_k^{(t)}-\frac{1}{m}\sum_{k=1}^m\widetilde{\bw}_{k}^{(t)}\|_2\leq 1$\footnote{It is a mild assumption since $\scal$ and $\scal^{(i)}$ differ by only one data point. The assumption is solely for the purpose of making the result more concise.}, then
\begin{align*}
        &\ebb_{\scal,A}\big[F(\overline{\bw}^{(t)})\unaryminus F_\scal(\overline{\bw}^{(t)})\big]\nonumber\\
        &\leq L_{\alpha,1}\big\{\frac{1}{mN}\sum_{k=1}^{m}\sum_{i=1}^{N}\ebb_{\scal,\scal^{(i)},A}\big[\|\bw_{k}^{(t)}\scalebox{0.8}{-}\widetilde{\bw}_k^{(t)}\|_2^2\big]\big\}^{\frac{1\scalebox{0.6}{+}\alpha}{2}}\scalebox{0.8}{+} \phi_\scal,
\end{align*}
where $\overline{\bw}^{(t)}= \frac{1}{m}\sum_{k=1}^m \bw_{k}^{(t)}$ represents the global averaged model, $L_{\alpha,1}=\frac{L}{1+\alpha}+\frac{1}{2}$ is a constant, and $\phi_\scal = \ebb_{\scal,A}\big[\frac{1}{2N}\sum_{i=1}^{N}\|\nabla f(\overline{\bw}^{(t)};z_i)\|^2_2\big]$ denotes the empirical gradient norm.
\end{lemma}
We give the proof in \cref{sec:pf-gen-stab}.

\cref{prop:gen-stab} suggests that if the consensus model learned by the distributed SGD $A$ is $\epsilon$-stable in the sense of \cref{def:dis-aver-stab}, the generalization error of the consensus model is bounded by $(\frac{L}{1\scalebox{0.6}{+}\alpha}\scalebox{0.8}{+}\frac{1}{2})\epsilon^{\frac{1\scalebox{0.8}{+}\alpha}{2}}\scalebox{0.8}{+}\phi_\scal$. 
The last term $\phi_\scal$ is very small for over-parameterized models near local or global minima \citep{vaswani2019fast}.
\cref{prop:gen-stab} improves Theorem 2 (c) of \citet{pmlr-v119-lei20c} by removing the  $\mathcal{O}{(\ebb_{S,A}\big[F^{\frac{2\alpha}{1+\alpha}}(A(S))\big])}$ term, where $\ebb_{S,A}\big[F(A(S))\big]$ denotes the population risk of the learned model $A(\scal)$ (see \cref{inq:refined-lei}). This improvement is significant, because $\ebb_{S,A}\big[F(A(S))\big]$ usually does not converge to zero in practice.

We now prove the generalization bound of D-SGD based on \cref{thm:on-average-holder} and \cref{prop:gen-stab}.

\begin{theorem}[Generalization Bound in Expectation with $\eta_t\equiv\eta$]\label{thm:gen-expected} Let all the assumptions of \cref{thm:on-average-holder} hold. With a fixed step sizes of $\eta_t\equiv\eta\leq 2L (1\unaryminus\frac{2}{m})$, the generalization error of the consensus model learned by D-SGD can be controlled as
\begin{align*}
        &\ebb_{\scal,A}\big[F(\overline{\bw}^{(t)})\unaryminus F_\scal(\overline{\bw}^{(t)})\big]\nonumber\\
        &\leq   L_{\alpha,2}
        \{\mathcal{O}{((\frac{\epsilon_{\scal}}{N})^{\frac{1\scalebox{0.6}{+}\alpha}{2}}\unaryplus \underbrace{{((1\unaryminus \frac{1}{m})\boldsymbol{\lambda^2}\scalebox{0.8}{+} \frac{1}{m})}^{\frac{1\scalebox{0.6}{+}\alpha}{2}}}_{{\text {Error from decentralization}}})}\}\scalebox{0.8}{+} \phi_\scal,
\end{align*}
where $\overline{\bw}^{(t)}= \frac{1}{m}\sum_{k=1}^m \bw_{k}^{(t)}$ represents the global averaged model, $L_{\alpha,2}=(\frac{L}{1+\alpha}\unaryplus\frac{1}{2})/{[1\unaryminus2\eta L (1\unaryminus\frac{1}{n})]}^{\frac{1+\alpha}{2}} $ is a constant, $\phi_\scal = \ebb_{\scal,A}\big[\frac{1}{2N}\sum_{i=1}^{N}\|\nabla f(\overline{\bw}^{(t)};z_i)\|^2_2\big]$ denotes the empirical gradient norm, and $\epsilon_{\scal}$ is the upper bounder of $ \frac{1}{m}\sum_{k=1}^{m}\ebb_{\scal,A}\big[F_{\scal}^{\frac{2\alpha}{1+\alpha}}(\bw_{k}^{(t)})\big]$.
\end{theorem}

The order of the generalization bound in \cref{thm:gen-expected} is $\mathcal{O}{(N^{-(1\unaryplus\alpha)/2}\unaryplus m^{-(1+\alpha)/2}\unaryplus\lambda^{1+\alpha} \unaryplus \phi_\scal)}$ and becomes $\mathcal{O}{(N^{-1}\unaryplus m^{-1}\unaryplus\lambda^2 \unaryplus \phi_\scal)}$ in the smooth settings where $\alpha=1$. The proof is provided in \cref{sec:pf-gen-stab}.

\begin{remark}
    \cref{th:stab-fix-eta} and \cref{thm:gen-expected} indicate that the stability and generalization of D-SGD are positively related to the spectral gap $1-\lambda$.
        The intuition of the results is that D-SGD with a denser connection topology (i.e., larger $\lambda$) can aggregate more information from its neighbors, thus ``indirectly" accessing more data at each iteration, leading to better generalization.
\end{remark}

\subsection{Practical Implications}
Our theory delivers significant practical implications.

\textbf{Communication topology and generalization}.
The intensity of communication is controlled by the spectral gap $1-\lambda$ of the underlying communication topologies (see \cref{tab:spectral-gap}). Detailed analyses of the spectral gaps of some commonly-used topologies can be found in Proposition 5 of \citet{nedic2018network} and \citet{ying2021exponential}.
Substituting the spectral gap of different topologies in \cref{tab:spectral-gap} into \cref{thm:gen-expected}, we can conclude that the generalization error of different topologies can be ranked as follows: fully-connected $<$ exponential $<$ grid $<$ ring, since 
\begin{multline*}
        0 < 1-\mathcal{O}{((\log _{2}(m))^{-1})}\\ < 1-\mathcal{O}{((m\log _{2}(m))^{-1})} < 1-\mathcal{O}{(m^{-2})}.
\end{multline*}
\begin{table}[!t]
  \centering
  \caption{Spectral gap of gossip matrices with different topology.\\}
    \scalebox{1.0}{
    \begin{tabular}{l|c}
    \toprule
    \toprule
    \textbf{Graph topoloy} & \textbf{Spectral gap} $1-\lambda$ \\
    \midrule
    \textbf{Disconnected} & 0 \\
     \textbf{Ring} & $\mathcal{O}{(1/m^2)}$  \\
    \textbf{Grid} &  $\mathcal{O}{(1/(m\log _{2}(m)))}$\\
     \textbf{Exponential} &  $ \mathcal{O}{({1}/{\log _{2}(m))}}$\\
    \textbf{Fully-connected} & 1 \\
    \bottomrule
    \bottomrule
    \end{tabular}%
    }
  \label{tab:spectral-gap}%
\end{table}%
On the one hand, our theory provides theoretical evidence that D-SGD generalizes better on well-connected topologies (i.e., topologies with larger spectral gap). On the other hand, we prove that for a specific topology, the worker number impacts the generalization of D-SGD through affecting the spectral gap of the topology. 

\textbf{Consensus distance control}. Recently, a line of studies have been devoted to understanding the connection between optimization and generalization through studying the effect of early phase training \citep{large-batch,achille2018critical,Frankle2020The}. In the decentralized settings, \citet{pmlr-v139-kong21a} claims that there exists a “critical consensus distance” in the initial training phase---consensus distance (i.e., $\frac{1}{m} \sum_{i=1}^{m}\|\bw_{k}^{(t)}-\frac{1}{m}\sum_{k=1}^m \bw_{k}^{(t)}\|^{2}_F$) below the critical threshold ensures good generalization. However, the reason why consensus distance control can promote generalization remains an open problem. Fortunately, the following theorem can explain this phenomenon by connecting the consensus distance notion in \citet{pmlr-v139-kong21a} with the algorithmic stability and the generalizability of D-SGD. 

\begin{corollary}
    \label{thm:consensus-control} Let all the assumptions of \cref{thm:on-average-holder} plus \cref{ass:lipschitz} and \cref{ass:l-smooth} hold. Suppose that the consensus distance satisfies $ \Gamma^2\leq\frac{1}{m}\sum_{k=1}^{m}\|\bw_{k}^{(\tau)}-{\overline{\bw}^{(\tau)}}\|_{2}^{2}\leq K^2$ for $\tau\leq t_\Gamma$, and is controlled below $\Gamma^2$ for $\tau>t_\Gamma$. We can conclude that the distributed on-average stability bound of D-SGD increases monotonically with $t_\Gamma$, if the total number of iterations $t\geq \frac{-C}{2\ln{C}}$.
\end{corollary}

We give the proof in \cref{sec:pf-imp}.

\cref{thm:consensus-control} provides theoretical evidence for the following empirical findings: (1) consensus control is beneficial for the algorithmic stability and thus for the generalizability of D-SGD; and (2) it is more effective to control the consensus distance at the initial stage of training than at the end of training.

%% file: section/5-empirical_results.tex
\section{Empirical Results}\label{sec:empirical results}
This section empirically validates our theoretical results. 
We first introduce the experimental setup and then study how the communication topology and the worker number affect the generalization of D-SGD. The code is available at \url{https://github.com/Raiden-Zhu/Generalization-of-DSGD}.

\begin{figure*}[t!]
\begin{subfigure}[VGG-11 on CIFAR-10, 32 workers]{.319\textwidth}
   \centering
  \includegraphics[width=1.05\linewidth]{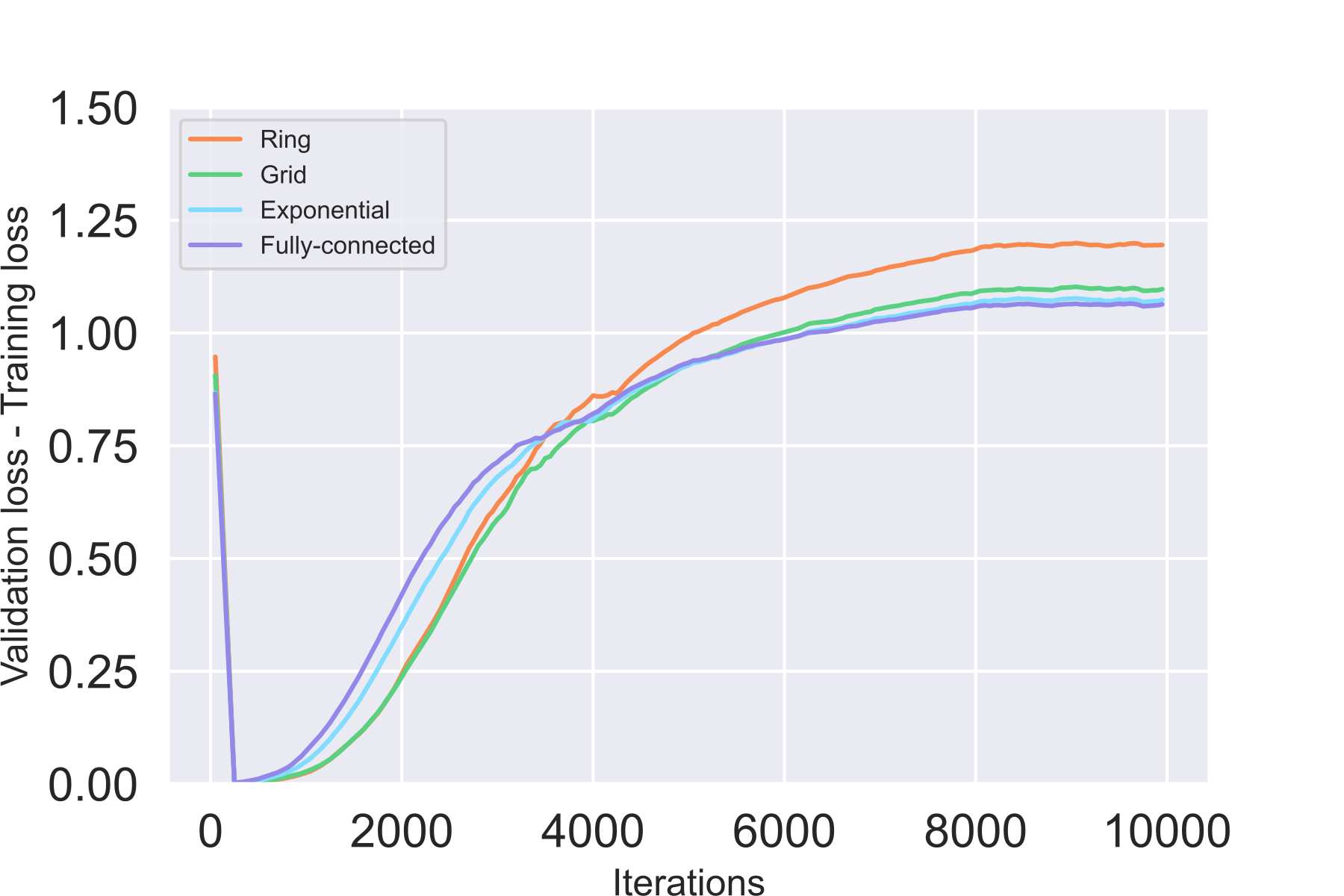}  
  \caption{VGG-11 on CIFAR-10, 32 workers}
\end{subfigure}\hfil
\begin{subfigure}[VGG-11 on CIFAR-100, 32 workers]{.319\textwidth}
  \centering
  \includegraphics[width=1.05\linewidth]{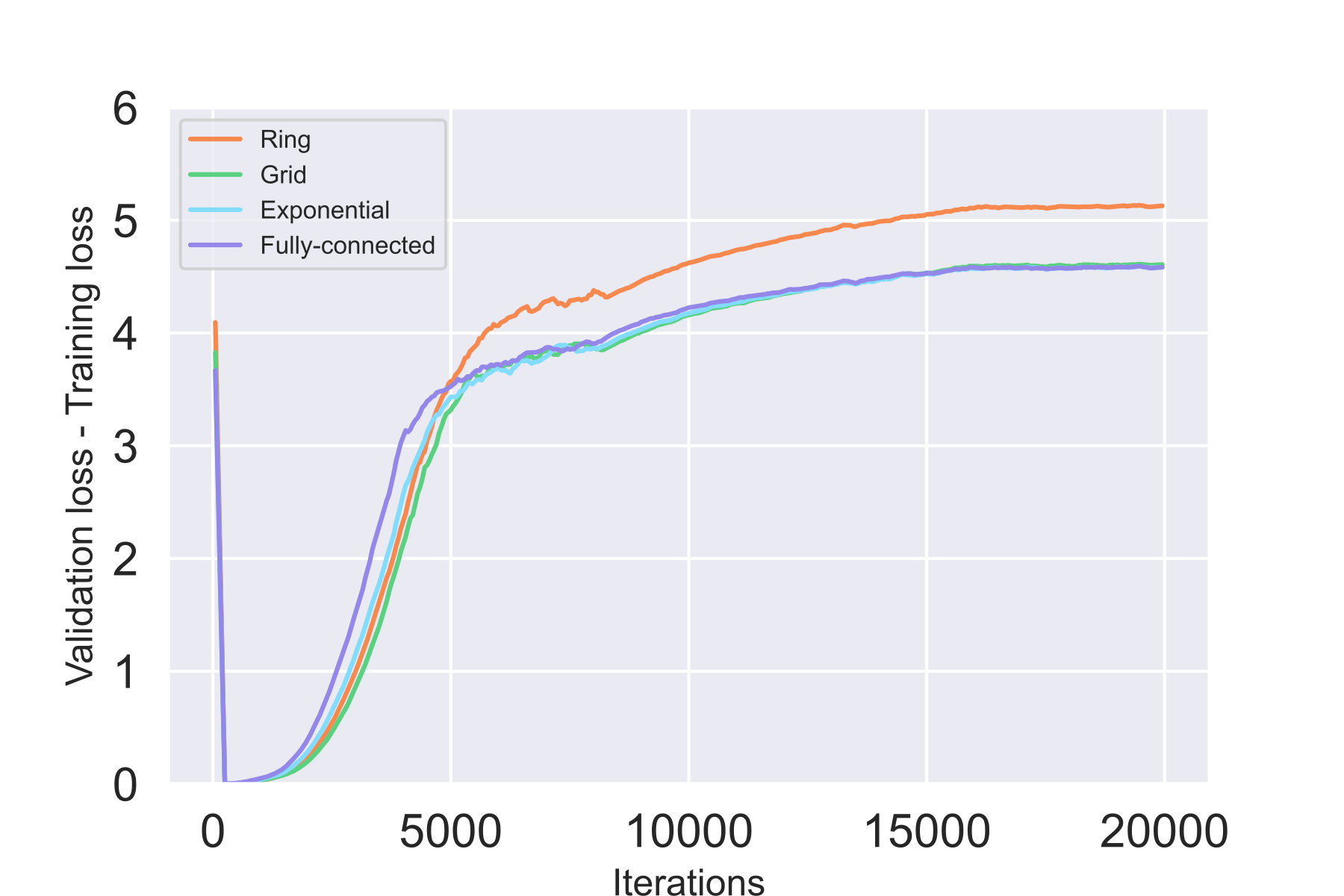}  
  \caption{VGG-11 on CIFAR-100, 32 workers}
\end{subfigure}\hfil
\begin{subfigure}[VGG-11 on Tiny ImageNet, 32 workers]{.319\textwidth}
  \centering
  \includegraphics[width=1.05\linewidth]{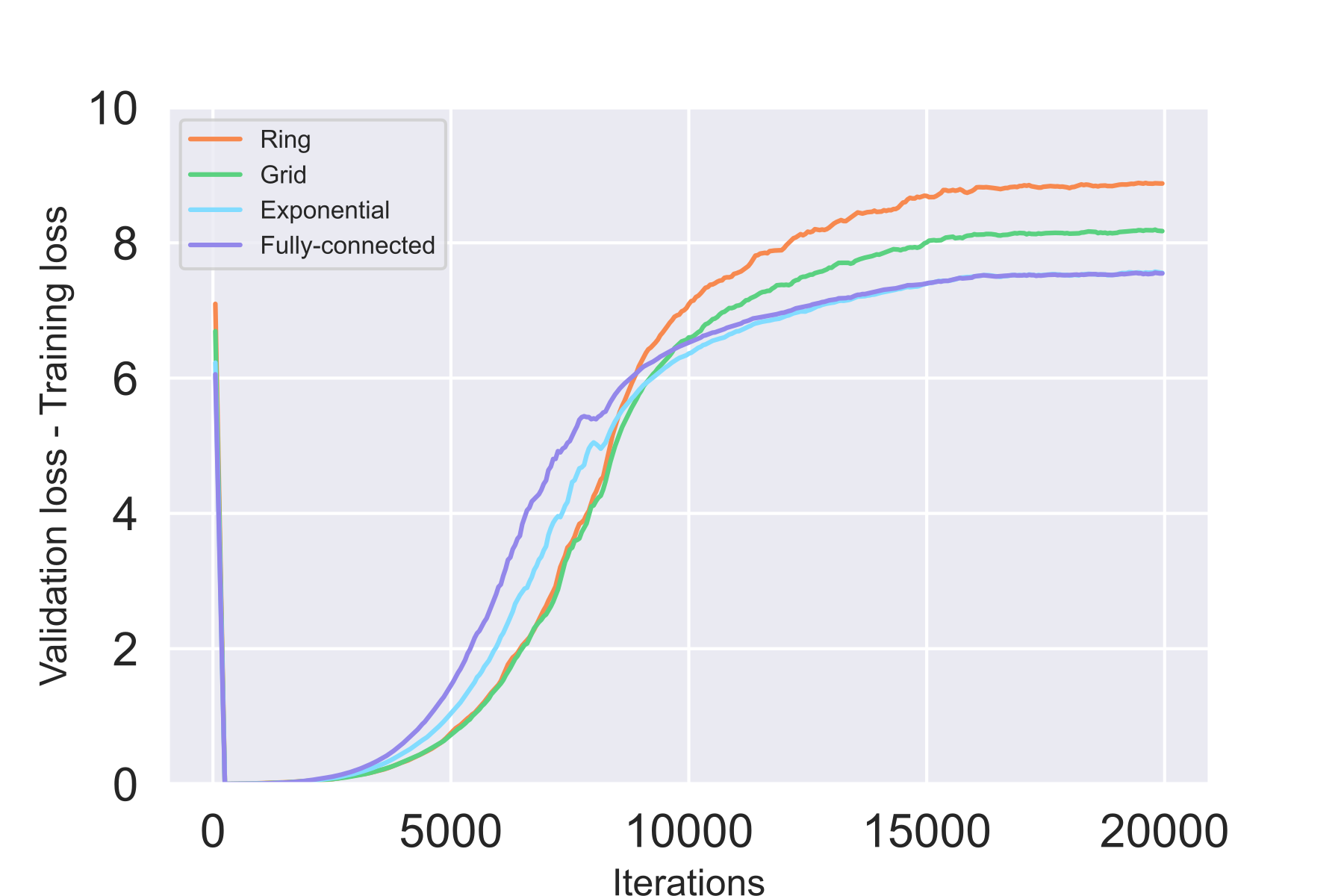}  
  \caption{VGG-11 on Tiny ImageNet, 32 workers}
\end{subfigure}

\medskip
\begin{subfigure}[VGG-11 on CIFAR-10, 64 workers]{.319\textwidth}
   \centering
  \includegraphics[width=1.05\linewidth]{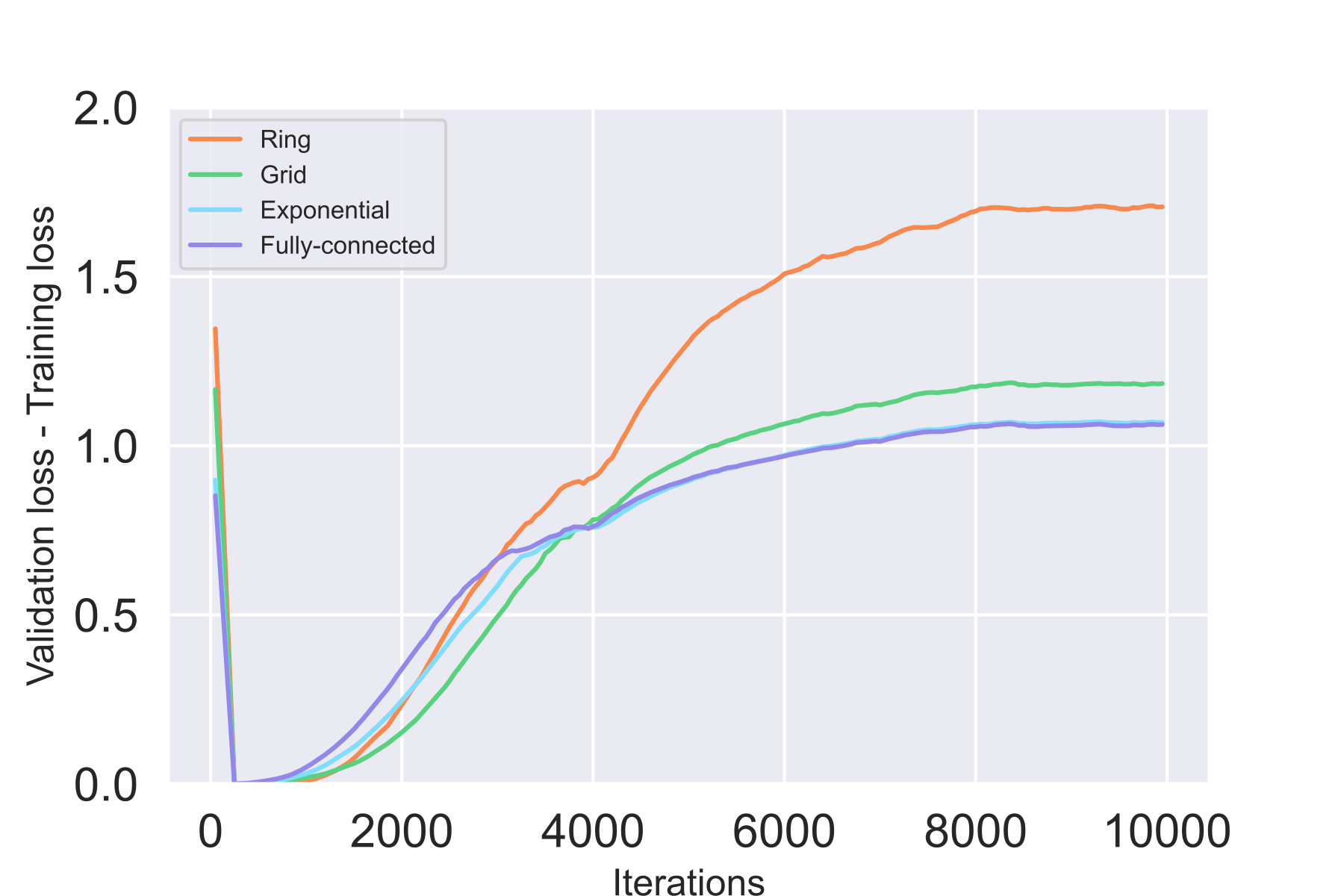}  
  \caption{VGG-11 on CIFAR-10, 64 workers}
\end{subfigure}\hfil
\begin{subfigure}[VGG-11 on CIFAR-100, 32 workers]{.319\textwidth}
  \centering
  \includegraphics[width=1.05\linewidth]{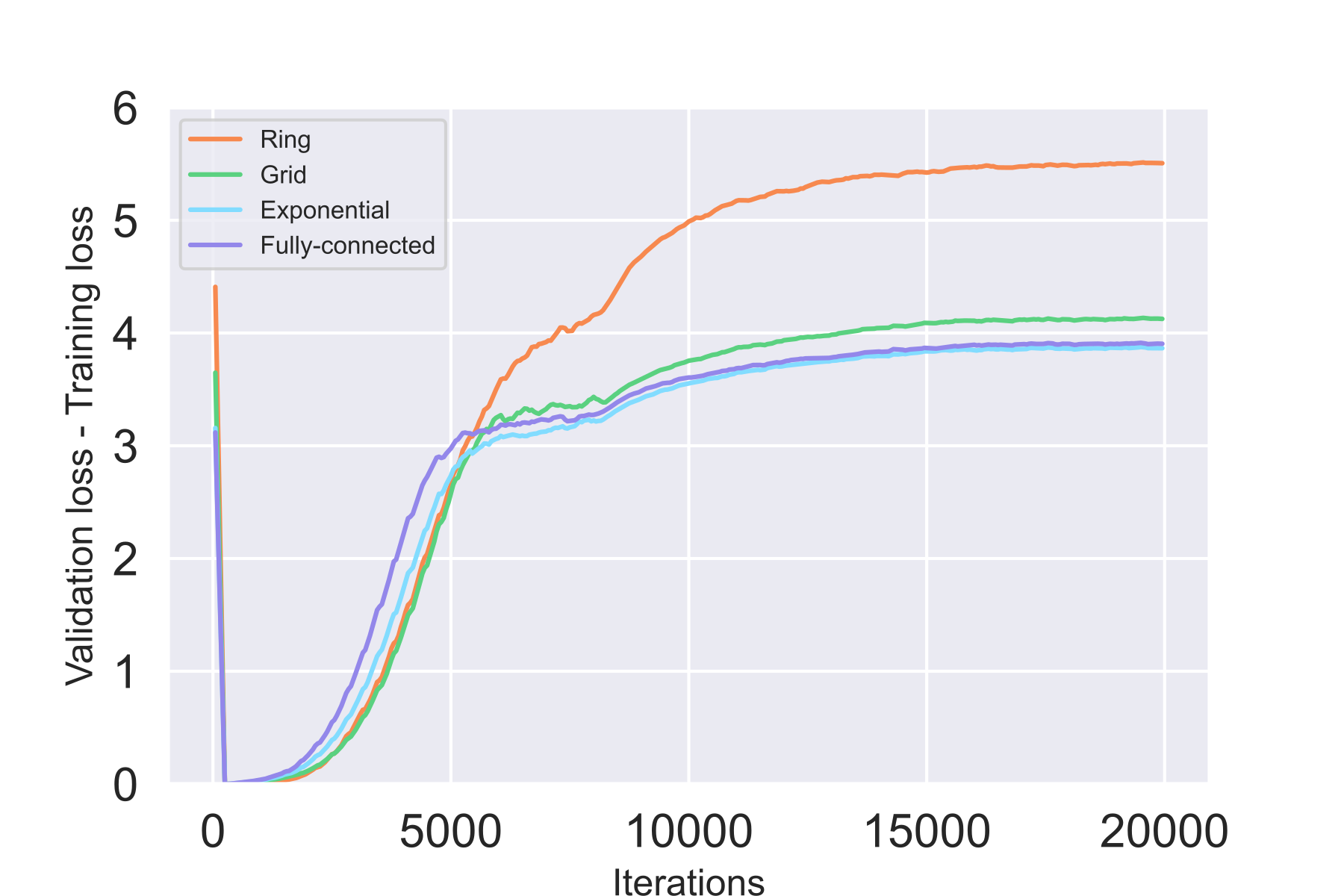}  
  \caption{VGG-11 on CIFAR-100, 64 workers}
\end{subfigure}\hfil
\begin{subfigure}[VGG-11 on Tiny ImageNet, 64 workers]{.319\textwidth}
  \centering
  \includegraphics[width=1.05\linewidth]{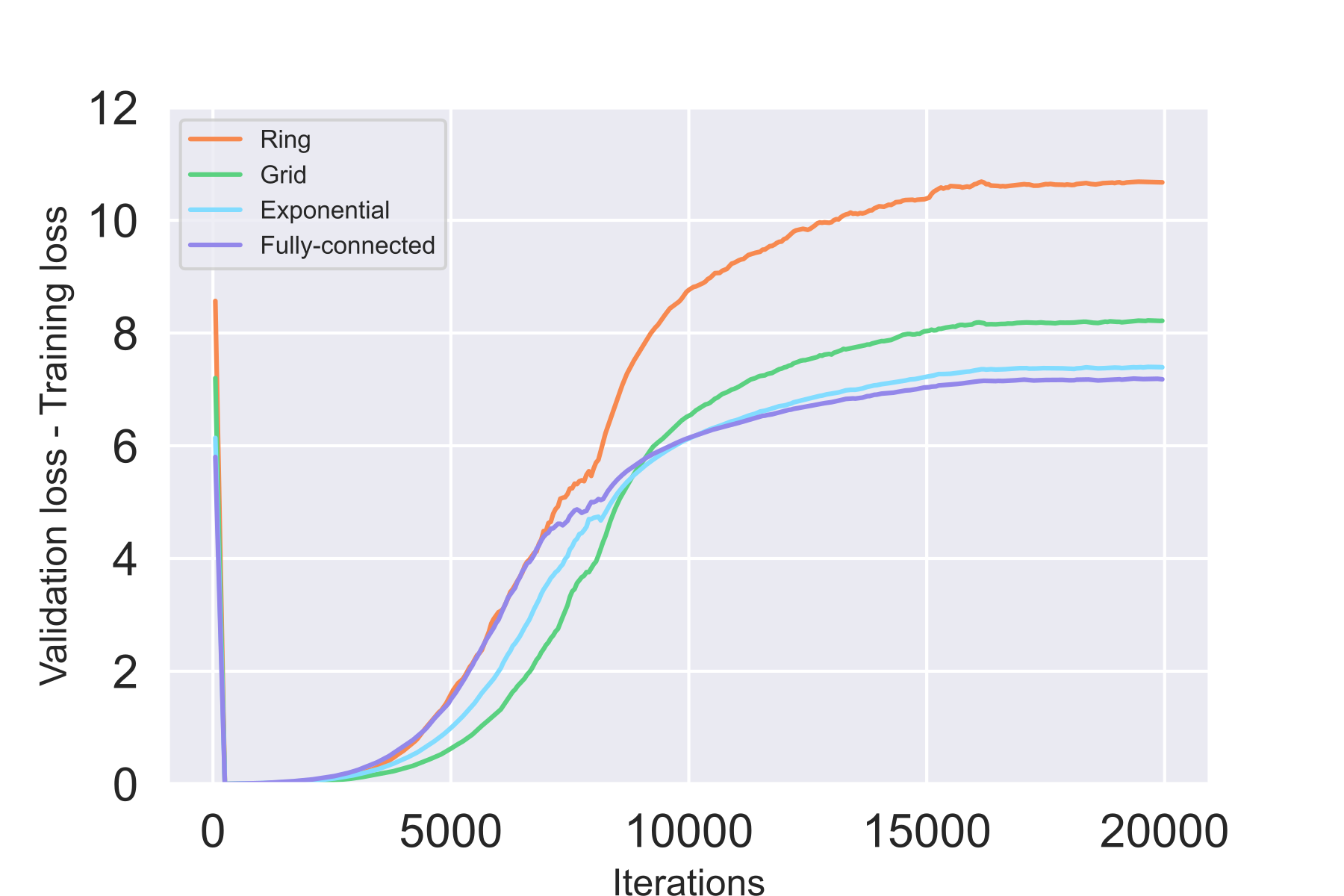}  
  \caption{VGG-11 on Tiny ImageNet, 64 workers}
\end{subfigure}
\caption{Loss differences of training VGG-11 with D-SGD on different topologies.}
\label{fig:topology-comparistion}
\end{figure*}

\subsection{Experimental Setup}

\textbf{Networks and datasets.} Network architectures VGG-11 \citep{simonyan2014very} and ResNet-18 \citep{he2016identity} are employed in our experiments. The models are trained on CIFAR-10, CIFAR-100 \citep{krizhevsky2009learning} and Tiny ImageNet \citep{le2015tiny}, three popular benchmark image classification datasets. The CIFAR-10 dataset consists of 60,000 32$\times$32 color images across 10 classes, with each class containing 5,000 training and 1,000 testing images. The CIFAR-100 dataset also consists of 60,000 32$\times$32 color images, except that it has 100 classes, each class containing 5,00 training and 1,00 testing images. Tiny ImageNet contains 120,000 64$\times$ 64 color images in 200 classes, each class containing 500 training images, 50 validation images, and 50 test images. No other pre-processing methods are employed.

\textbf{Training setting.} Vanilla D-SGD is employed to train image classifiers based on VGG-11 and ResNet-18 on fully-connected, ring, grid, and static exponential topologies. The number of workers is set as 32 and 64. 
Batch normalization \citep{ioffe2015batch} and dropout \citep{JMLR:v15:srivastava14a} are employed in training VGG-11. The local batch size is set as 64. To control the impact of different total batch size (local batch size $\times$ worker number) caused by the different number of workers, we apply the linear scaling law (i.e., linearly increase learning rate w.r.t. total batch size) \citep{he2016deep,goyal2017accurate}. The initial learning rate is set as $0.4$ and will be divided by 
$10$ when the model has accessed $2/5$ and $4/5$ of the total number of iterations \citep{he2016deep}.
All other techniques, including momentum \citep{qian1999momentum}, weight decay \citep{tihonov1963solution}, early  stopping \citep{Bengio2012} and data augmentation \citep{lecun1998gradient} are disabled.

\textbf{Implementations.} All our experiments are conducted on a computing cluster with GPUs of NVIDIA\textsuperscript{\textregistered} Tesla\textsuperscript{\texttrademark} V100 16GB and CPUs of Intel\textsuperscript{\textregistered} Xeon\textsuperscript{\textregistered} Gold 6140 CPU @ 2.30GHz.
Our code is implemented based on PyTorch \citep{paszke2019pytorch}.

\subsection{Communication topology and generalization}

We calculate 
the difference between the validation loss and the training loss on different topologies separately, as shown in \cref{fig:topology-comparistion} and \cref{fig:topology-comparistion-resnet}. Two observations are obtained from those figures: (1) for large topologies, the loss differences can be sorted as follows: fully-connected $\approx$ exponential $<$ grid $<$ ring; (2) as the worker number increases, the loss differences of D-SGD on different topologies increase. These observations suggest that (1) D-SGD generalizes better on well-connected topology with a larger spectral gap; (2) the generalizability gap of D-SGD on different topologies grows as the worker number increases.

In comparison to ResNet-18, we observe when VGG-11 is chosen as the backbone, the generalization gaps between different topologies are larger (see \cref{fig:topology-comparistion} and \cref{fig:topology-comparistion-resnet}). We conjecture that the generalization gaps are amplified because the H\"older constant $L$ of VGG-11 is larger than that of ResNet-18, according to our theory suggesting that the generalization error of D-SGD linearly increases with $L$ (see \cref{thm:gen-expected}).
 
One may observe that the loss difference of the fully-connected topology is larger than that of other topologies in the initial training phase, which seems to be inconsistent with our theory.
 Theoretically, explaining this phenomenon is an optimization problem, which is beyond the scope of this work.
 One possible explanation is motivated by the stability-convergence trade-off in iterative optimization algorithms \citep{chen2018stability}. Since the fully-connected system converges faster, the corresponding optimization error is smaller than the other three kinds of systems at the beginning of training. Therefore, the fully-connected system is less stable in the initial phase of training.

%% file: section/6-discussion.tex
\section{Future Works}\label{sec:discussion}



\textbf{Implicit bias of D-SGD.} As pointed out in \citet{zhang2021loss}, the additional gradient noise in D-SGD helps it converge to a flatter minima compared to centralized distributed SGD.
Therefore, a direct question is whether there is a superior implicit bias effect \citep{soudry2018implicit,ji2018gradient,NEURIPS2019_dbc4d84b,wang2021implicit} in D-SGD compared to centralized distributed SGD, which involves the convergence direction?
Would decentralization change the implicit bias?
Can we derive fine-grained generalization bounds of D-SGD based on the implicit bias analysis? 

%% file: section/7-conclusion.tex
\section{Conclusion}

In this paper, we theoretically analyze the algorithmic stability and generalizability of decentralized stochastic gradient descent (D-SGD).
We prove that the consensus model learned by D-SGD is $\mathcal{O}{(N^{-1} + m^{-1} +\lambda^2)}$-stable, where $N$ is the total sample size on each worker, $m$ is the worker number, and $1-\lambda$ is the spectral gap of the communication topology. Based on this stability result, we obtain an $\mathcal{O}{(N^{-(1+\alpha)/2}+ m^{-(1+\alpha)/2}+\lambda^{1+\alpha} + \phi_\scal)}$ in-average generalization bound, characterizing the gap between the training performance and the test performance.
Our error bounds are non-vacuous, even when the worker
number is sufficiently large, or the communication topology is sufficiently sparse.
According to our theory, we can conclude: (1) the generalizability of D-SGD is positively correlated with the spectral gap of the underlying communication topology; (2) the generalizability of D-SGD decreases when the worker number increases. These theoretical findings are then empirically justified by the experiments of VGG-11 and ResNet-18 on CIFAR-10, CIFAR-100, and Tiny ImageNet. 
The theory can also explain why consensus control at the beginning of training is able to promote the generalizability of D-SGD.

%% file: section/acknowledgement
\newpage
\section*{Acknowledgement}

This work is supported by the Major Science and Technology Innovation 2030 key projects ``New Generation Artificial Intelligence'' (No. 2021ZD0111700), the National Natural Science Foundation of China (No. 61976186, U20B2066, and 61932016), the Key Research and Development Program of Zhejiang Province (No. 2021C01164), the Fundamental Research Funds for the Central Universities (No. 226-2022-00064 and WK2150110024), and the Starry Night Science Fund of Zhejiang University Shanghai Institute for Advanced Study (No. SN-ZJU-SIAS-001). This work was completed when Tongtian Zhu and Zhengyang Niu were interns at JD Explore Academy.

The authors would like to thank Chun Li, Yingjie Wang, Haowen Chen, and Shaopeng Fu for their insightful comments on the revision of this manuscript and Li Shen, Rong Dai, and Luofeng Liao for their helpful discussions. We appreciate Batiste Le Bars and Xiaolin Hu for pointing out issues in Lemma 3, which have been addressed. We also sincerely thank the anonymous ICML reviewers and chairs for their constructive comments.

%% file: section/appendix.tex
\bibliography{example_paper}
\bibliographystyle{icml2022}

\newpage
\onecolumn
\appendix
\numberwithin{equation}{section}
\numberwithin{theorem}{section}
\numberwithin{remark}{section}
\numberwithin{definition}{section}
\numberwithin{assumption}{section}
\numberwithin{figure}{section}
\numberwithin{table}{section}
\renewcommand{\thesection}{{\Alph{section}}}
\renewcommand{\thesubsection}{\Alph{section}.\arabic{subsection}}
\renewcommand{\thesubsubsection}{\Alph{section}.\arabic{subsection}.\arabic{subsubsection}}

\newpage

\section{Additional Background}\label{sec:add-back-ground}

The following remarks clarify some notations in the literature.
\begin{remark}
Stochastic learning algorithms $A: \cup_{n} \mathcal{Z}^{n} \mapsto \mathcal{W}$ are often applied to produce an output model $A(\scal)\in\rbb^d$ based on the training set $\scal$. To avoid ambiguity, let $A(\scal)$ denote the model generated by a general learning algorithm, and $\bw$ denote the models generated by a specific stochastic learning algorithm. 
\end{remark}

\begin{remark}
Stochastic learning introduces two kinds of randomness: one from the sampling of training examples and another from the adopted randomized algorithm. In the following analysis, $\ebb_A[\cdot]$ stands for the expectation w.r.t. the randomness of the algorithm $A$, and $\ebb_S[\cdot]$ denotes the expectation w.r.t. the randomness originating from sampling the data set $\scal$. Notice that $\scal \sim \mathcal{D}^n$ and $z \sim \mathcal{D}$, therefore $\ebb_\scal[\cdot]$ differs from $\ebb_z[\cdot]$ defined in \cref{eq:risk}.
\end{remark}

Based on the work by \citet{pmlr-v48-hardt16} and \citet{NIPS2007_0d3180d6}, we give a formulation of excess error decomposition and demonstrate how to understand generalization through error decomposition.
\begin{definition}[Excess Error Decomposition\label{def:ex-error}]
We denote the empirical risk minimization (ERM) solution by $\bw_{\scal}^*=\arg\min_{\bw}F_{\scal}(\bw)$ and $\bw^*=\arg\min_{\bw}F(\bw)$. The excess error $F(A(\scal))-F(\bw^*)$ can be decomposed as
\begin{multline}\label{eq:serial-(d+1)ecomposition}
\underbrace{\mathbb{E}_{\scal, A}\left[F(A(\scal))-F\left(\bw^{*}\right)\right]}_{\text {Excess error }}\\
=\underbrace{\mathbb{E}_{\scal, A}\left[F(A(\scal))-F_{\scal}(A(\scal))\right]}_{\text {Generalization error}}+\underbrace{\mathbb{E}_{\scal, A}\left[F_{\scal}(A(\scal))-F_{\scal}\left(\bw_{\scal}^{*}\right)\right]}_{{\text {Optimization error }}}+\mathbb{E}_{\scal, A}\left[F_{\scal}(\bw_{\scal}^{*})-F\left(\bw^{*}\right)\right]\\
\leq \underbrace{\mathbb{E}_{\scal, A}\left[F(A(\scal))-F_{\scal}(A(\scal))\right]}_{\text {Generalization error}}+\underbrace{\mathbb{E}_{\scal, A}\left[F_{\scal}(A(\scal))-F_{\scal}\left(\bw_{\scal}^{*}\right)\right]}_{{\text {Optimization error }}}
\end{multline}
The last inequality holds since and $\mathbb{E}_{\scal, A}\left[F_{\scal}(\bw_{\scal}^{*})\right]\leq\mathbb{E}_{\scal, A}\left[F_\scal\left(\bw^{*}\right)\right] = \mathbb{E}_{\scal}\big[F(\bw^*)$.
The empirical risk and the population risk above are defined in \cref{eq:risk}.
This paper considers upper bounding the first term called the generalization error.
\end{definition}
Lipschitzness and smoothness are two commonly adopted assumptions to establish the uniform stability guarantees of SGD.
\begin{assumption}[Lipschitzness\label{ass:lipschitz}]
  $\big\|\nabla f(\bw;z)\big\|_2\leq G$ for all $\bw\in\rbb^d$ and $z\in\zcal$. 
\end{assumption}
\begin{assumption}[Smoothness\label{ass:l-smooth}]
$f$ is $\beta$-smooth if for any $z$ and $\bw, \widetilde{\bw}\in\rbb^d$,
  \begin{equation}
    \big\|\nabla f(\bw;z)-\nabla f(\widetilde{\bw};z)\big\|_2\leq \beta\|\bw-\widetilde{\bw}\|_2.
  \end{equation}
\end{assumption}
These two restrictive assumptions are not satisfied in many real contexts. For example, the Lipschitz constant $G$ can be very large for some learning problems \citep{NEURIPS2019_95e1533e,pmlr-v119-lei20c}. In addition, neural nets with piecewise linear activation functions like ReLU are not smooth. Smoothness is generally difficult to ensure at the beginning and intermediate phases of deep neural network training \citep{NEURIPS2020_2e2c4bf7}. 

\begin{assumption}[H\"older Continuity\label{def:holder}]
  Let $L>0,\alpha\in[0,1]$. $\nabla f(\cdot,z)$ is $(\alpha,L)$-H\"older continuous if for all $\bw,\widetilde{\bw}\in\rbb^d$ and $z\in \mathcal{Z}$,
  \begin{equation}\label[inequality]{holder-condition}
    \big\|\nabla f(\bw;z)-\nabla f(\widetilde{\bw};z)\big\|_2\leq L\|\bw-\widetilde{\bw}\|_2^\alpha.
  \end{equation}
\end{assumption}

H\"older continuous gradient assumption is much weaker than smoothness by definition. Serving as an intermediate class of functions $(C^{1, \alpha}\left(\mathbb{R}^{n}\right))$ between smooth functions $(C^{1, 1}\left(\mathbb{R}^{n}\right))$ and functions with Lipschitz continuous gradients $(C^{1, 0}\left(\mathbb{R}^{n}\right))$, the main advantage of functions with  H\"older continuous gradients lies in the ability to automatically adjust the smoothness parameter to a proper level \citep{nesterov2015universal}: \cref{holder-condition} with $\alpha=1$ corresponds to smoothness and 
\cref{holder-condition} with $\alpha=0$ is equivalent to Lipschitzness (see \cref{ass:lipschitz}).

\begin{assumption}[Gaussian Weight Difference\label{ass:gaussian-diff-w}] We assume that the difference between $\bw_k^{(t)}$ and $\widetilde{\bw}_k^{(t)}$ (the $t$-th iterate on $k$-th worker produced by \cref{eq:dec-sgd-entry} based on $\scal_k$ and $\scal_k^{(i)}$  respectively) is independent and normally distributed:

\[
    \big(\bw_k^{(t)}-\widetilde{\bw}_k^{(t)}\big) {\sim} \mathcal{N}(\mu_{t,k}, \sigma_{t,k}^2 I_{d}),\quad k = 1, \dots, m
\]

where $I_{d}$ denotes an identity matrix with size $d$, and $\mu_{t,k}, \sigma_{t,k}^2$ are unknown parameters. We also give a mild constraint that the $d$-dimensional parameter $\mu_{t,k}$ satisfies $ d\cdot\mu_0^2\leq\|\mu_{t,k}\|_2^2\leq d\cdot\mu^2$ and the parameter $\sigma_{t,k}^2\in\rbb$ is bounded by $\sigma^2$. \cref{ass:gaussian-diff-w} is mild since $\scal_k$ and $\scal_k^{(i)}$ only vary at one point.
\end{assumption}

Commonly used stability notions are listed below.
\begin{definition}[Hypothesis Stability\label{def:hyp-stab}]
  A stochastic algorithm $A$ is hypothesis $\epsilon$-stable w.r.t. the loss function $f$ if for all training data sets $\scal,\scal^{(i)}\in\zcal^n$ that differ by at most one example, we have
  \begin{equation}\label{hyp-stab}
  \ebb_z\ebb_A\big[f(A(\scal);z)-f(A(\scal^{(i)});z)\big]\leq\epsilon.
  \end{equation}
\end{definition}

\begin{definition}[Uniform Stability\label{def:unif-stab}]
  A stochastic algorithm $A$ is $\epsilon$-uniformly stable w.r.t. the loss function $f$ if for all training data sets $\scal,\scal^{(i)}\in\zcal^n$ that differ by at most one example, we have
  \begin{equation}\label{unif-stab}
  \sup_z\ebb_A\big[f(A(\scal);z)-f(A(\scal^{(i)});z)\big]\leq\epsilon.
  \end{equation}
\end{definition}

\begin{definition}[On-average Model Stability, \citep{pmlr-v119-lei20c}\label{def:aver-stab}]
  A stochastic algorithm $A$ is $\ell_2$ on-average model $\epsilon$-stable for all training data sets $\scal,\scal^{(i)}\in\zcal^n$ that differ by at most one example, we have
  \[
   \frac{1}{N}\sum_{i=1}^{N}\ebb_{\scal,\scal^{(i)},A}\big[\|A(\scal)-A(\scal^{(i)})\|_2^2\big]\leq\epsilon^2.
  \]
\end{definition}
Some widely used notions regarding decentralized training are listed as follows.
\begin{definition}[Doubly Stochastic Matrix\label{def:dou-matrix}]
  Let $\mathcal{G}=(\mathcal{V},\mathcal{E})$ stand for the decentralized communication topology where $\mathcal{V}$ denotes the set of m computational nodes and $\mathcal{E}$ represents the edge set. For any given graph $\mathcal{G}=(\mathcal{V},\mathcal{E})$,
the doubly stochastic gossip matrix ${\bp} = [\bp_{k,l}] \in \mathbb{R}^{m\times m}$ is defined on the edge set $\mathcal{E}$ that satisfies:
(1) If $k\neq l$ and $(k,l) \notin {\cal E}$, then $\bp_{k,l} =0$ (disconnected); otherwise, $\bp_{k,l} >0$ (connected);
(2) $\bp_{k,l}\in [0,1]\ \forall k, l$;
(3) ${\bp} = {\bp}^{\top}$;
and (4) $\sum_k \bp_{k,l}=\sum_l \bp_{k,l}=1$ (standard weight matrix for undirected graph).
\end{definition}
\begin{definition}[Spectral Gap\label{def:spectral-gap}]
Denote $\lambda=\max \left\{\left|\lambda_{2}\right|, \ldots, \left|\lambda_{i}\right|, \ldots, \left|\lambda_{m}\right|\right\}$ where $\lambda_i$ denotes the $i$-th largest eigenvalue of gossip matrix ${\bp}\in\rbb^{m\times m}$.
  The spectral gap of a gossip matrix $\bp$ can be defined as follows:
  \begin{equation*}
  {\text spectral\ gap} := 1- \lambda.
  \end{equation*}
According to the definition of doubly stochastic matrix (\cref{def:dou-matrix}), we have $0\leq\lambda<1$. The spectral gap measures the connectivity of the communication topology, which is close to 0 for sparse topologies and will approach 1 for well-connected topologies.
\end{definition}

To facilitate our subsequent analysis, we provide some preliminaries of matrix algebra here.
\begin{definition}[Frobenius Norm\label{def:Frobenius norm}]
  The Frobenius norm (Euclidean norm, or Hilbert–Schmidt norm) is the matrix norm of a matrix $\mathbf{A}\in\rbb^{p\times q}$ defined as the square root of the sum of the squares of its elements:
  \begin{equation*}
\|\mathbf{A}\|_{F} = \sqrt{\sum_{i=1}^{q} \sum_{j=1}^{q}\left|a_{i j}\right|^{2}}=\sqrt{\Tr(\mathbf{A}\tran \mathbf{A})}.
    \end{equation*}
    For any $\mathbf{A},\mathbf{B}\in\rbb^{p\times q}$, the following identity holds:
\[
    \|\mathbf{A}+\mathbf{B}\|_{\text{F}}^{2}=\|\mathbf{A}\|_{\text{F}}^{2}+\|\mathbf{B}\|_{\text{F}}^{2}+2\langle \mathbf{A},\mathbf{B}\rangle_F,
\]
where $\langle \cdot,\cdot\rangle _{\text{F}}$ is the Frobenius inner product.
\end{definition}

\newpage
\section{Additional Related Work}\label{sec:add-related-work}

\subsection{Non-centralized learning.}
To handle an increasing amount of data and model parameters, distributed learning across multiple computing nodes (workers) emerges. A traditional distributed learning system usually follows a centralized setup \citep{abadi2016tensorflow}. However, such a central server-based learning scheme suffers from two main issues: (1) A centralized communication protocol significantly slows down the training since central servers are easily overloaded, especially in low-bandwidth or high-latency cases \citep{NIPS2017_f7552665}; (2) There exists potential information leakage through privacy attacks on model parameters despite decentralizing data using federated learning \citep{NEURIPS2019_60a6c400,NEURIPS2020_c4ede56b,Yin_2021_CVPR}. As an alternative, training in  a  non-centralized fashion allows workers to  balance the load on the central server through the gossip technique, as well as  maintain confidentiality \citep{warnat2021swarm}. 
Model decentralization can be divided into three kinds of categories by layers \citep{pmlr-v139-lu21a}: (1) On the application layer, decentralized training usually refers to federated learning \citep{zhao2018federated,dai2022dispfl} ; (2) On the protocol layer, decentralization denotes average gossip where local workers communicate by averaging their parameters with their neighbors on a graph \citep{NIPS2017_f7552665} and (3) on the topology layer, it means a sparse topology graph \citep{wan2020rat}. 

\subsection{Generalization via algorithmic stability.}

Algorithmic stability theory, PAC-Bayes theory, and information theory are major tools for constructing algorithm-dependent generalization bounds \citep{pmlr-v134-neu21a}.
A direct intuition behind algorithmic stability is that if an algorithm does not rely excessively on any single data point, it can generalize well. Proving generalization bounds based on the sensitivity of the algorithm to changes in the learning sample can be traced back to \citet{vapnik1974theory} and \citet{devroye1979distribution}. After that, the celebrated work by \citet{bousquet2002stability} establishes the relationship between uniform stability and generalization in high probability. Follow-up work by \citet{shalev2010learnability} identifies stability as the major necessary and sufficient condition for learnability. Then, \citet{pmlr-v48-hardt16} provide uniform stability bounds for stochastic gradient methods (SGM) and show the strong stability properties of SGD with convex and smooth losses. Recent work by \citet{pmlr-v119-lei20c} defines a new on-average stability notion and conducts generalization analyses on SGD with the H\"older continuous assumption. In addition to uniform stability, there are other stability notions including on-average stability \citep{shalev2010learnability}, uniform argument stability \citep{liu2017algorithmic}, data-dependent stability \citep{pmlr-v80-kuzborskij18a}, hypothesis set stability \citep{NEURIPS2019_300d1539} and locally elastic stability \citep{pmlr-v139-deng21b}.

\newpage
\section{Additional Experimental Results}
We also calculate the difference between the validation loss and
the training loss of training ResNet-18 with D-SGD.

\begin{figure*}[h!]
\begin{subfigure}[CIFAR-10, 32 workers]{.31\textwidth}
   \centering
  \includegraphics[width=1.1\linewidth]{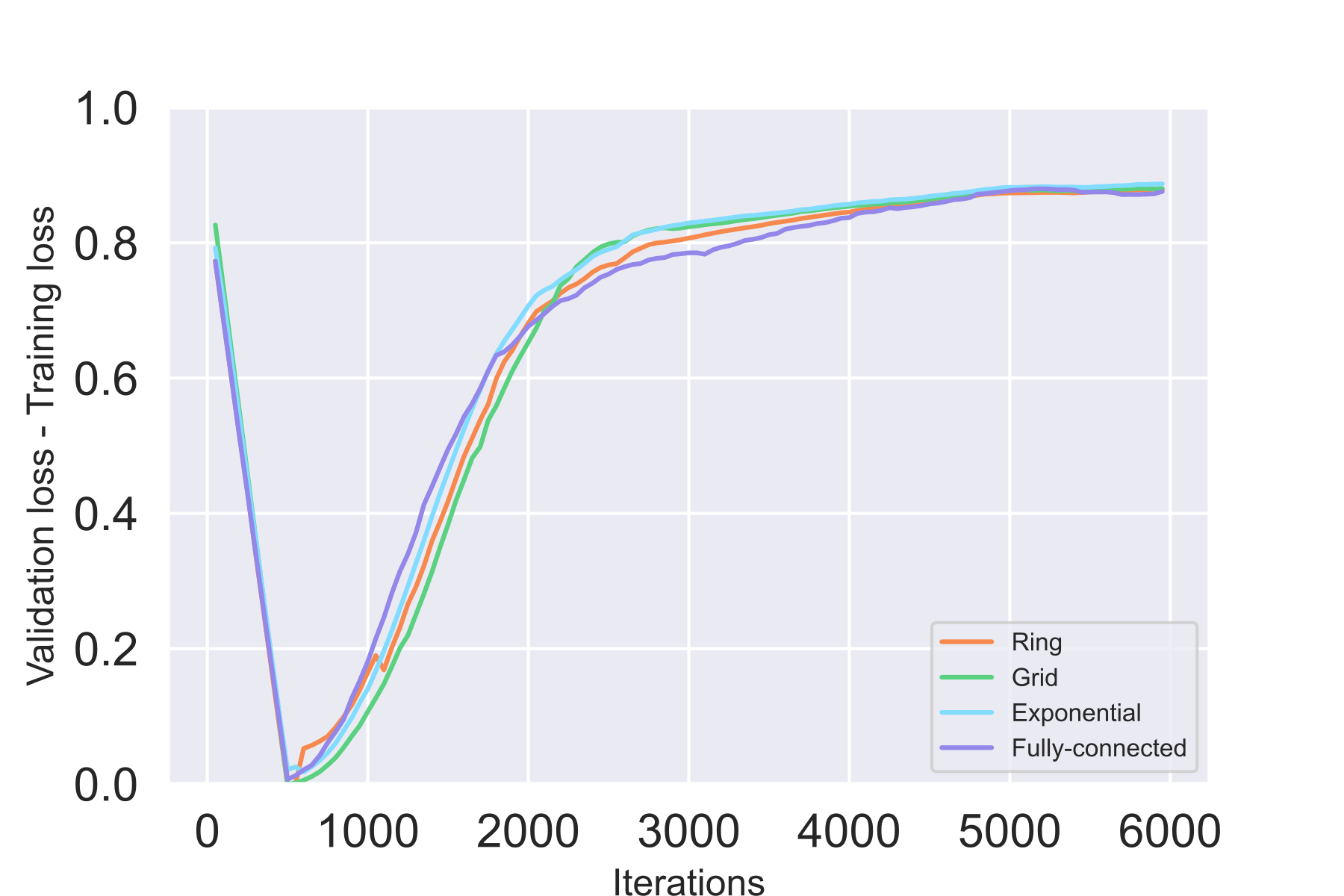}  
  \caption{CIFAR-10, 32 workers}
\end{subfigure}\hfil
\begin{subfigure}[CIFAR-100, 32 workers]{.31\textwidth}
  \centering
  \includegraphics[width=1.1\linewidth]{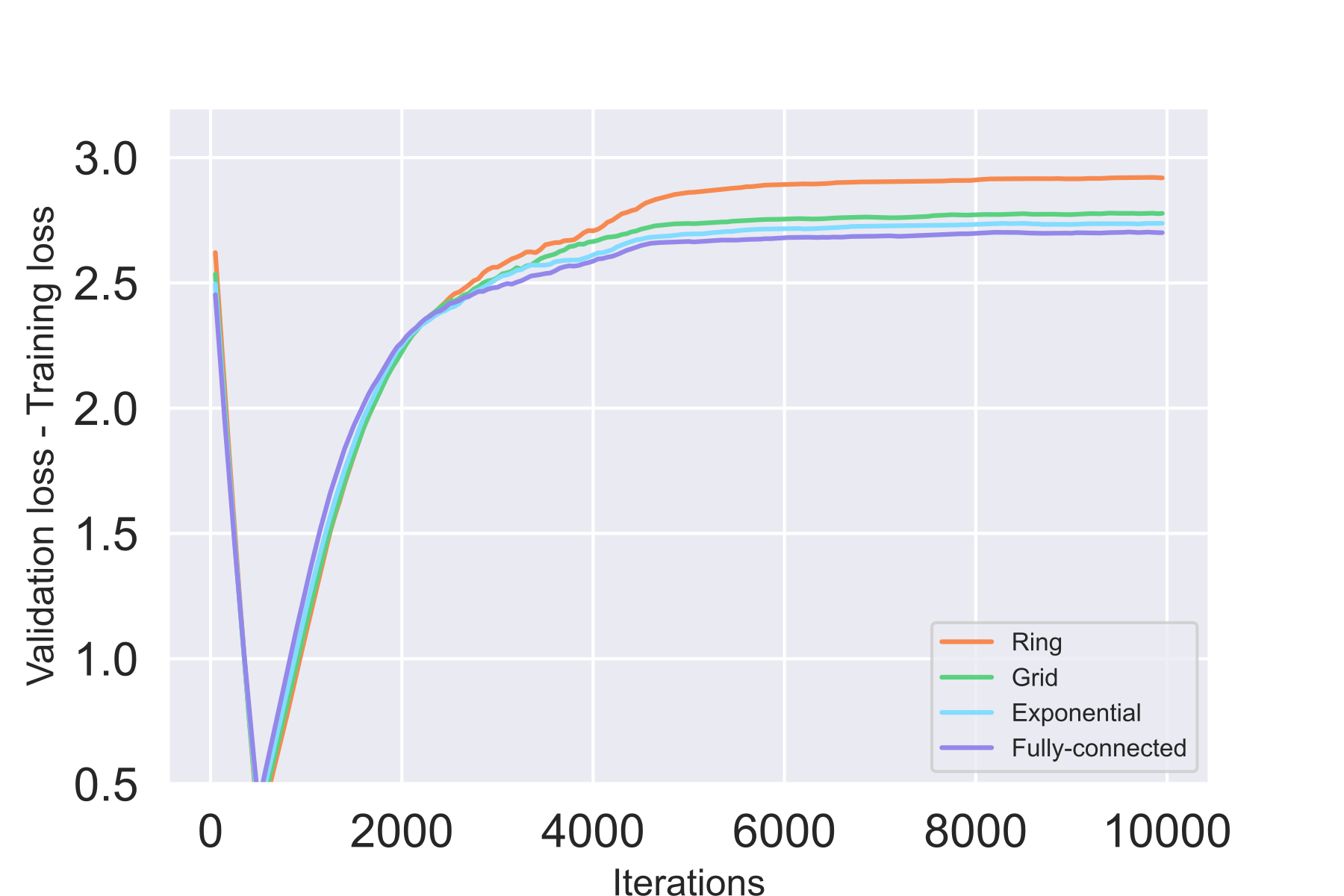}  
  \caption{CIFAR-100, 32 workers}
\end{subfigure}\hfil
\begin{subfigure}[Tiny ImageNet, 32 workers]{.31\textwidth}
  \centering
  \includegraphics[width=1.1\linewidth]{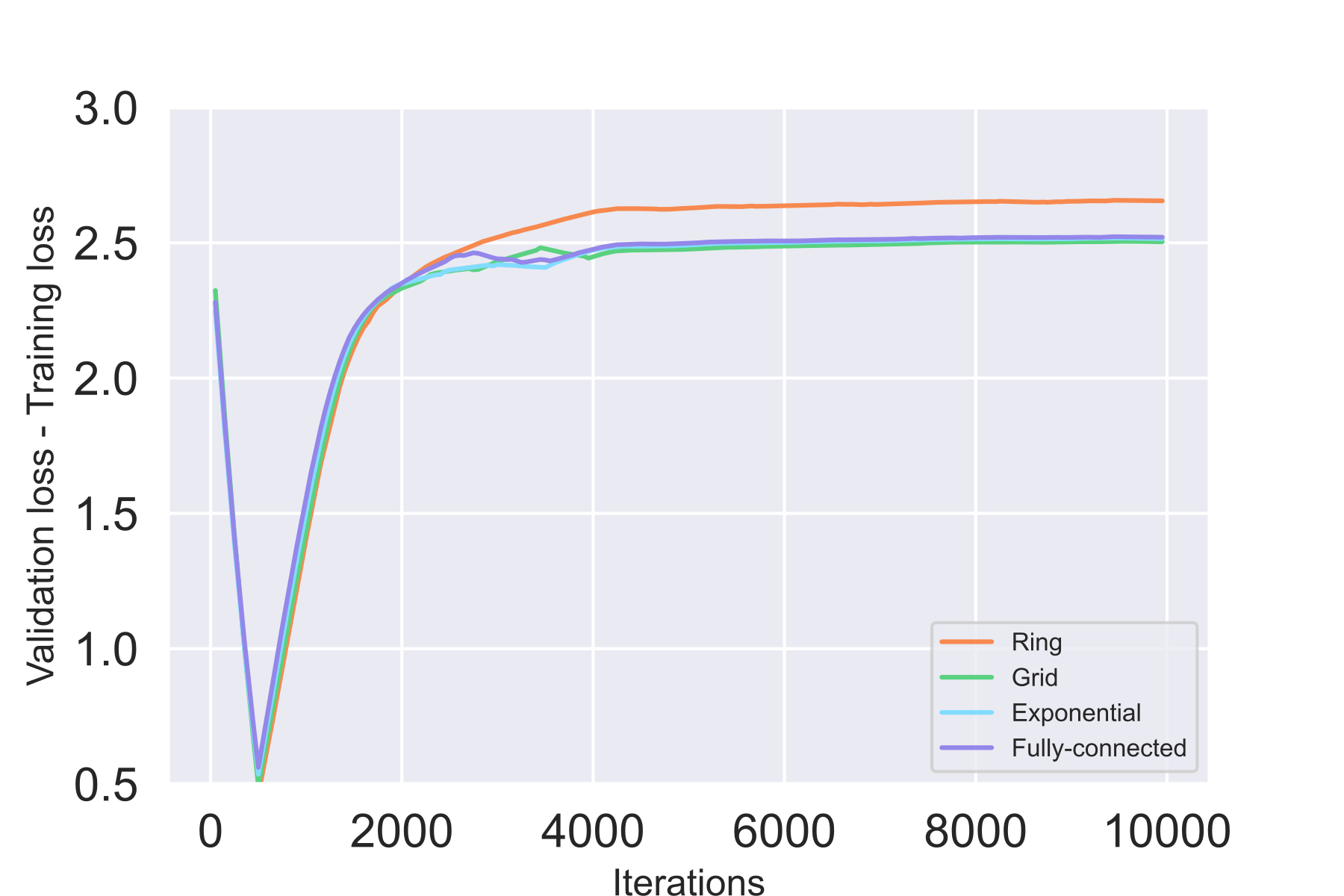}  
  \caption{Tiny ImageNet, 32 workers}
\end{subfigure}

\medskip
\begin{subfigure}[CIFAR-10, 64 workers]{.31\textwidth}
   \centering
  \includegraphics[width=1.1\linewidth]{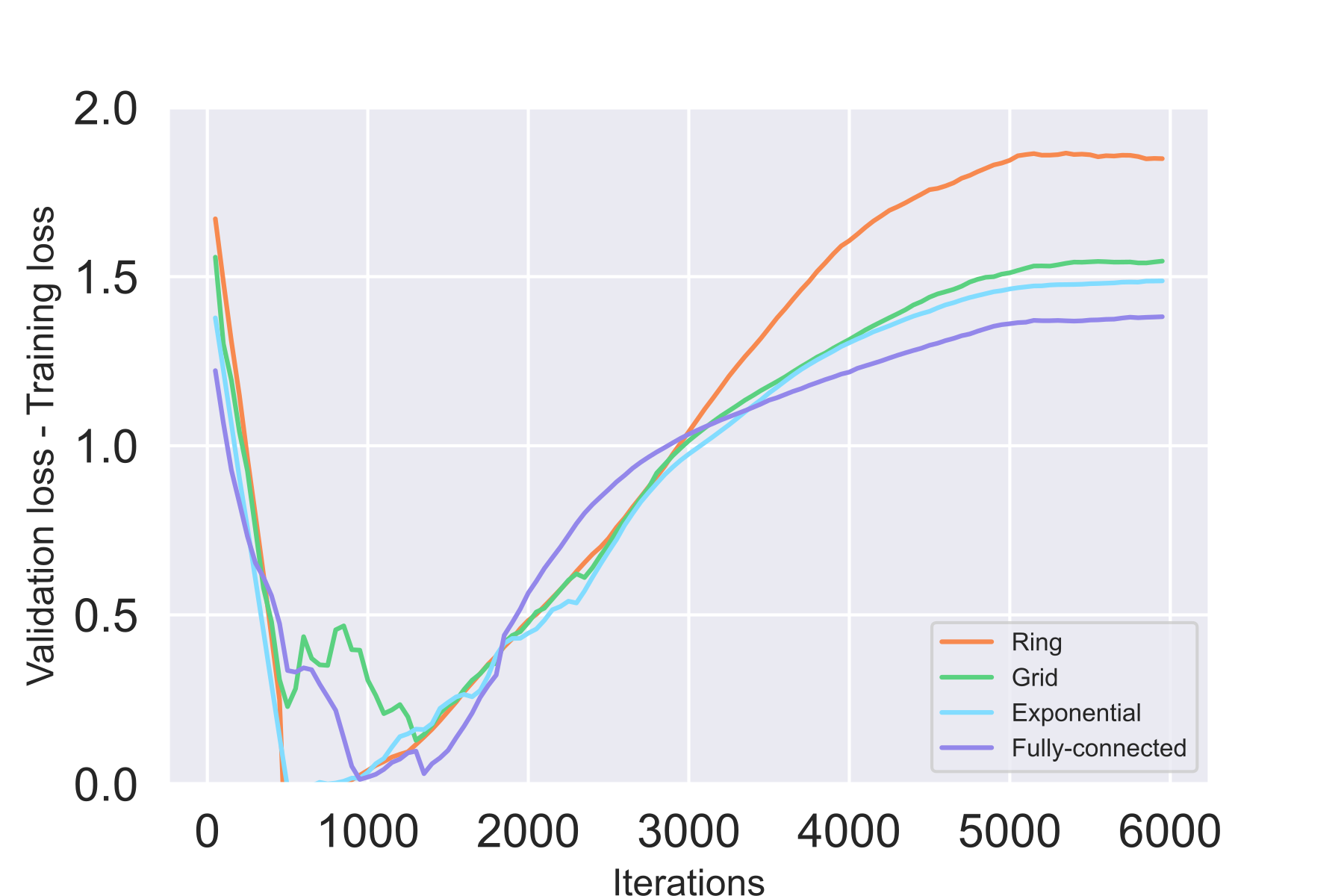}  
  \caption{CIFAR-10, 64 workers}
\end{subfigure}\hfil
\begin{subfigure}[CIFAR-100, 32 workers]{.31\textwidth}
  \centering
  \includegraphics[width=1.1\linewidth]{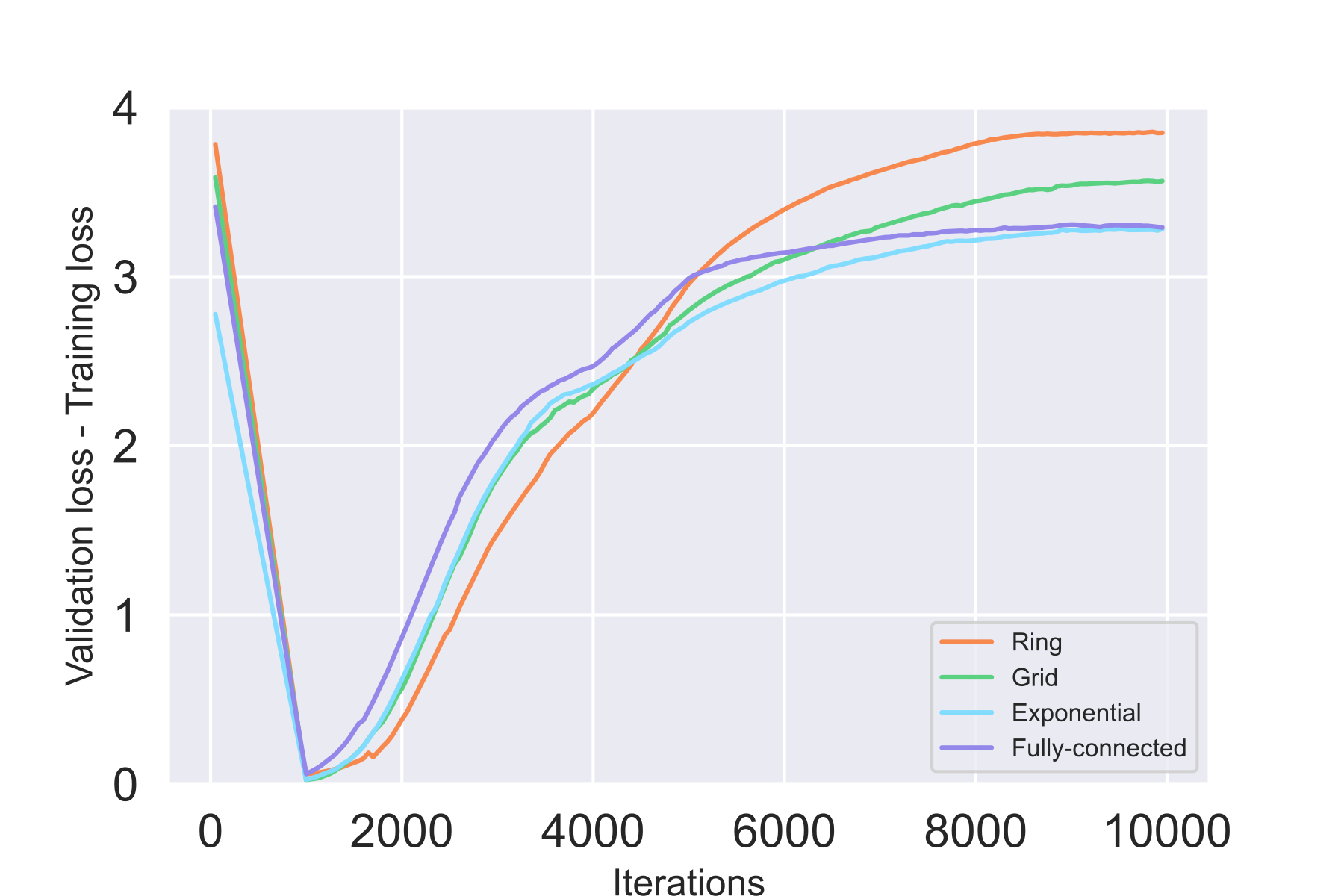}  
  \caption{CIFAR-100, 64 workers}
\end{subfigure}\hfil
\begin{subfigure}[Tiny ImageNet, 64 workers]{.31\textwidth}
  \centering
  \includegraphics[width=1.1\linewidth]{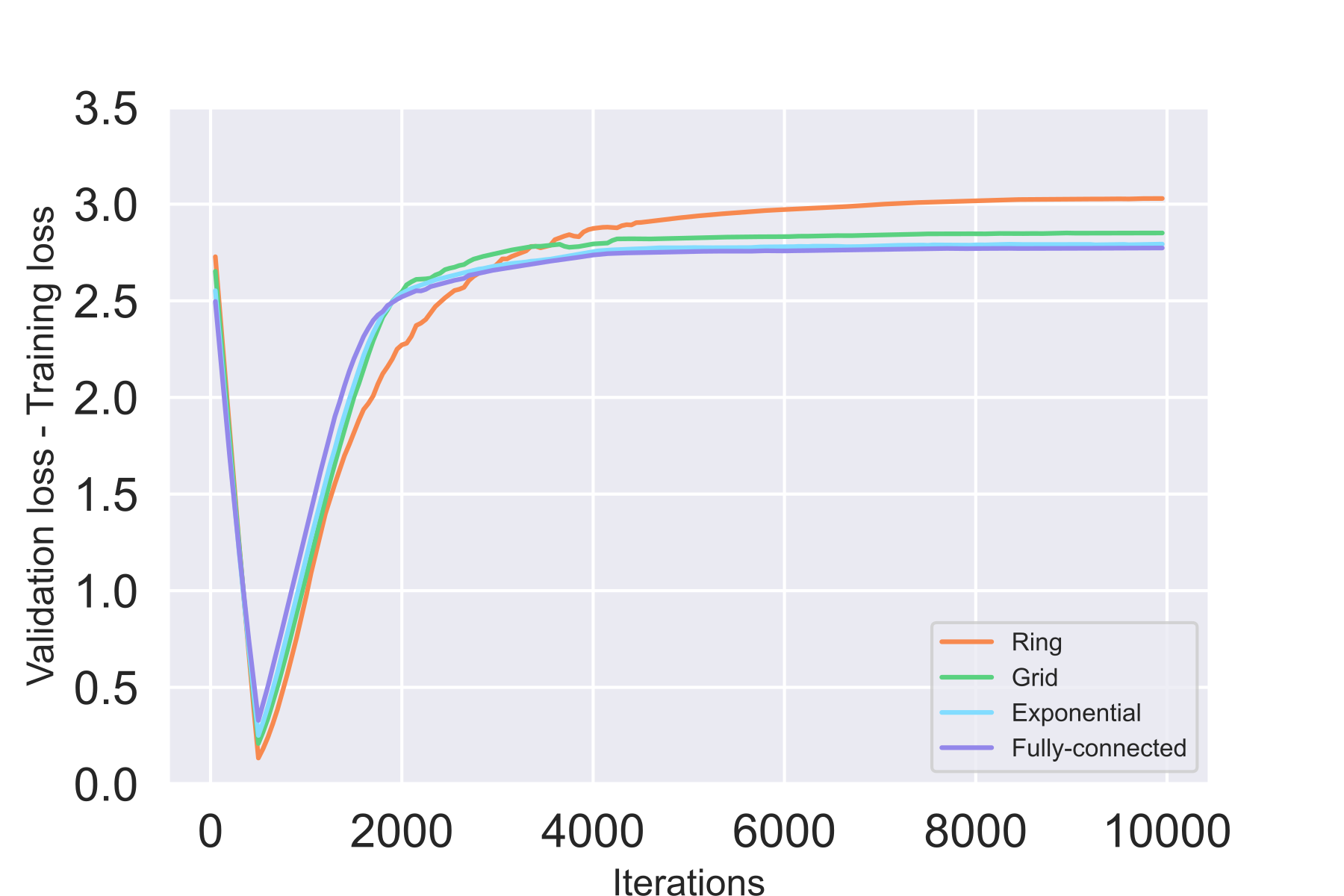}  
  \caption{Tiny ImageNet, 64 workers}
\end{subfigure}
\caption{Loss differences of training ResNet-18 with D-SGD on different topologies.}
\label{fig:topology-comparistion-resnet}
\end{figure*}

\newpage
\section{Proof}

\subsection{Technical lemmas\label{sec:lemmas}}

To complete our proof, we first introduce some technical lemmas.

\begin{lemma}[Corollary 1.14., \citep{TCS-003}\label{th:property-DS-matrix}]
    Let $\bM$ stand for the matrix with all the elements be $1/m$ and $\bp$ is defined in \cref{def:dou-matrix}. For any $k\in\mathbb{Z}^{+}$, the following inequality holds:
    \begin{equation}
        \left\|\bp^{k}-\bM\right\|_{2,2}\leq\left\|\bp\right\|^k_\lambda.
    \end{equation}
\end{lemma}

\begin{lemma}\label{th:standard-inequal}
    For any $a, b\in\rbb$ and $p\in\rbb^+$, the following inequality holds:
    \begin{equation}
        (a+b)^2\leq (1+p)a^2+(1+p^{-1})b^2.
    \end{equation}
\end{lemma}

\begin{lemma}[Self-bounding Property, \citep{pmlr-v119-lei20c}\label{lem:self-bounding}]
Assume that for all $z\in\zcal$, the map $\bw\mapsto f(\bw;z)$ is nonnegative with its gradient $\nabla f(\bw;z)$ being $(\alpha,L)$-H\"older continuous (\cref{def:holder}), then $\bw\mapsto f(\bw;z)$ can be bounded as
  \[
  \|\nabla f(\bw,z)\|_2\leq c_{\alpha,1}f^{\frac{\alpha}{1+\alpha}}(\bw,z),\quad\forall \bw\in\rbb^d,z\in\zcal.
  \]
  where
  \begin{equation}\label{alpha-1}
  c_{\alpha,1}=\begin{cases}
                 (1+1/\alpha)^{\frac{\alpha}{1+\alpha}}L^{\frac{1}{1+\alpha}}, & \mbox{if } \alpha>0 \\
                 \sup_z\|\nabla f(0;z)\|_2+L, & \mbox{if } \alpha=0.
               \end{cases}
\end{equation}
\end{lemma}

\begin{remark}
    The self-binding property implies that H"older continuous gradients can be controlled by function values. The $\alpha=1$ and $\alpha\in(0,1)$ case  are established by \citet{NIPS2010_76cf99d3} and \citet{ying2017unregularized}, respectively. The case where $\alpha=0$ follows directly from \cref{def:holder}.
\end{remark}


\begin{lemma}\label{inequal:simple-inequal}
    For any $a, b\in\rbb^d$ with $a_i$, $b_i$ being their $i$-th components, respectively, the following inequality holds:
    \begin{equation}
        a\tran b=\sum_{i}a_i b_i\leq \sqrt{\sum_i a_i^2 \sum_i b_i^2}\leq \frac{\sum_i a_i^2 + \sum_i b_i^2}{2}.
    \end{equation}
\end{lemma}

\begin{lemma}\label{lemma: upper-bounding-f(x)-f(y)}
    Let $x,y\in\rbb^d$, for any differentiable function $h\ ( \rbb^d \mapsto \rbb)$ which has $(\alpha,L)$-H\"older continuous gradient, we have
    \begin{equation}
        h(x) - h(y) \leq \frac{L}{1+\alpha} {\|x-y\|}_2^{1+\alpha} + \frac{1}{2}\|x-y\|^2_2 +
        \frac{1}{2}\|\nabla h(y)\|^2_2.
    \end{equation}
\end{lemma}

\textit{Proof of \cref{lemma: upper-bounding-f(x)-f(y)}.}

According to the definition of derivative, the difference between $h(x)$ and $h(y)$ can be written as
\begin{align*}
    h(x) - h(y) = \int_0^1 \langle \nabla h(u_{(t)}),x-y\rangle \mathrm{d}t,
\end{align*}
where $u_{(t)}=y+t(x-y).$

Subtracting both sides with $\langle \nabla h(u_{(t)}),x-y\rangle$ gives
\begin{align*}
    h(x) - h(y) - \langle \nabla h(y),x-y\rangle = \int_0^1 \langle \nabla h(u_{(t)})-\nabla h(y),x-y\rangle \mathrm{d}t.
\end{align*}
The triangle inequality and the $(\alpha,L)$-H\"older continuity of $h$ further guarantee
\begin{align*}
    |h(x) - h(y)| &= \int_0^1 |\langle \nabla h(u_{(t)})-\nabla h(y),x-y\rangle| \mathrm{d}t + |\langle \nabla h(y),x-y\rangle|\\
    & \leq \int_0^1 \| \nabla h(u_{(t)})-\nabla h(y)\| \| x-y\| \mathrm{d}t + |\langle \nabla h(y),x-y\rangle|\\
    & \leq \int_0^1 L\| u_{(t)}-y\|^\alpha \| x-y\| \mathrm{d}t + |\langle \nabla h(y),x-y\rangle|\\
    & \leq L \| x-y\|^{1+\alpha} \int_0^1 t^\alpha \mathrm{d}t + |\langle \nabla h(y),x-y\rangle|,
\end{align*}
Since $|\langle \nabla h(y), x-y\rangle|\leq \frac{1}{2}\|x-y\|^2_2 +
        \frac{1}{2}\|\nabla h(y)\|^2_2$, we arrive at
\begin{align*}
    h(x) - h(y) &\leq \frac{L}{1+\alpha} {\|x-y\|}_2^{1+\alpha} +
    |\langle \nabla h(y), x-y\rangle|\\
    &\leq \frac{L}{1+\alpha} {\|x-y\|}_2^{1+\alpha} + \frac{1}{2}\|x-y\|^2_2 +
        \frac{1}{2}\|\nabla h(y)\|^2_2.
\end{align*}



{\color{magenta}\qed}

\newpage
\subsection{Algorithmic stability of D-SGD\label{sec:proof-holder-stab}}

\textit{\textbf{Proof of \cref{thm:on-average-holder}.}}

To begin with, we decompose $\sum_{k=1}^{m}\big\|\bw_{k}^{(t+1)}-\widetilde{\bw}_{k}^{(t+1)}\big\|_{2}^{2}$, the on-average stability of D-SGD at the $t$-th iteration, into three parts by the definition of the vector 2-norm. In the following, we will let $z_{k,\zeta_t}^{(t)}$ and $\tilde{z}_{k,\zeta_t}^{(t)}$ denote two random data points drawn from $\scal_k$ and $\scal^{(i)}_k$ on the $k$-th worker at the $i$-th iteration, respectively\footnote{Note that $z_{k,\zeta_t}^{(t)}$ and $\tilde{z}_{k,\zeta_t}^{(t)}$ is the $\zeta_t$-th element of $\scal_k$ and $\scal^{(i)}_k$, respectively. According to the construction of $\scal_k$ and $\scal^{(i)}_k$ in \cref{def:dis-aver-stab}, we know that with probability $1-\frac{1}{N}$, $z_{k,\zeta_t}^{(t)}=\tilde{z}_{k,\zeta_t}^{(t)}$; and with probability $\frac{1}{N}$, $z_{k,\zeta_t}^{(t)}\neq\tilde{z}_{k,\zeta_t}^{(t)}$.}. 



\[
\begin{aligned}
\sum_{k=1}^{m}&\big\|\bw_{k}^{(t+1)}-\widetilde{\bw}_{k}^{(t+1)}\big\|_{2}^{2} \\
&=\sum_{k=1}^{m}\big\|\sum_{l=1}^{m} \mathbf{P}_{k, l} \bw_{t}(l)-\eta_{t} \nabla f\left(\bw_{k}^{(t)}; z_{k,\zeta_t}^{(t)}\right)-\sum_{l=1}^{m} \mathbf{P}_{k, l} \widetilde{\bw}_l^{(t)}+\eta_{t} \nabla f\left(\widetilde{\bw}_k^{(t)}, \tilde{z}_{k,\zeta_t}^{(t)}\right)\big\|_{2}^{2} \\
& =  \sum_{k=1}^{m}\big\|\sum_{l=1}^{m}\bp_{k,l} (\bw_l^{(t)}-\widetilde{\bw}_l^{(t)})\big\|_{2}^{2}
+\sum_{k=1}^{m}{\eta_t}^2\big\|\nabla f\left(\bw_{k}^{(t)}; z_{k,\zeta_t}^{(t)}\right)-\nabla f\left(\widetilde{\bw}_k^{(t)}, \tilde{z}_{k,\zeta_t}^{(t)}\right)\big\|_{2}^{2}\\
& -2\sum_{k=1}^{m} \eta_t\big\langle\sum_{l=1}^{m}\bp_{k,l} (\bw_l^{(t)}-\widetilde{\bw}_l^{(t)}), \nabla f\left(\bw_{k}^{(t)}; z_{k,\zeta_t}^{(t)}\right)-\nabla f\left(\widetilde{\bw}_k^{(t)}, \tilde{z}_{k,\zeta_t}^{(t)}\right)\big\rangle\\
&=\underbrace{\big\|\bp (\bW^{(t)}-\widetilde{\bW}^{(t)})\big\|_{F}^{2}}_{\text {$T_1$}}
 +\underbrace{\sum_{k=1}^{m} \ind_{z_{k,\zeta_t}^{(t)}\neq \tilde{z}_{k,\zeta_t}^{(t)}}  \big[\big\|\bw_{k}^{(t+1)}-\widetilde{\bw}_{k}^{(t+1)}\big\|_{2}^{2}-\big\|\sum_{l=1}^{m}\bp_{k,l} (\bw_l^{(t)}-\widetilde{\bw}_l^{(t)})\big\|_{2}^{2}\big]}_{\text {$T_2$}}\\
&+  \sum_{k=1}^{m} \ind_{z_{k,\zeta_t}^{(t)}= \tilde{z}_{k,\zeta_t}^{(t)}} \big[{\eta_t}^2\big\|\nabla f\left(\bw_{k}^{(t)}; z_{k,\zeta_t}^{(t)}\right)-\nabla f\left(\widetilde{\bw}_k^{(t)}, \tilde{z}_{k,\zeta_t}^{(t)}\right)\big\|_{2}^{2}\\
&\underbrace{\qquad\qquad\qquad\qquad\quad\quad\ \  -2\eta_t\big\langle\sum_{l=1}^{m}\bp_{k,l} (\bw_l^{(t)}-\widetilde{\bw}_l^{(t)}), \nabla f\left(\bw_{k}^{(t)}; z_{k,\zeta_t}^{(t)}\right)-\nabla f\left(\widetilde{\bw}_k^{(t)}, \tilde{z}_{k,\zeta_t}^{(t)}\right)\big\rangle\big]}_{\text {$T_3$}}
\end{aligned}
\]

where $\left\|\cdot\right\|_F$ denotes the Frobenius norm (see \cref{def:Frobenius norm}).\\

(1) To construct our proof, we start by constructing an upper bound of $\ebb_A (T_1)$:
\begin{equation}
    \ebb_A (T_1)=\ebb_A\big[\big\|\bp (\bW^{(t)}-\widetilde{\bW}^{(t)})\big\|_{F}^{2}
    \leq d(\sigma^2 + \mu^2)\big[(m-1)\lambda^2+1\big].\label[inequality]{ineq:T_1-th1_1}
\end{equation}
\textit{Proof.}

The on-averaged stability after a single gossip communication can be written as
\begin{align}
    \big\|\bp (\bW^{(t)}-\widetilde{\bW}^{(t)})\big\|_{F}^{2}
    & = \sum_{k=1}^{m}\big\|\sum_{l=1}^{m}\bp_{k,l} (\bw_l^{(t)}-\widetilde{\bw}_l^{(t)})\big\|_{2}^{2} 
    = \sum_{v=1}^{d} \sum_{k=1}^{m}\big\|\sum_{l=1}^{m}\bp_{k,l} (\bw_l^{v,(t)}-\widetilde{\bw}_l^{v,(t)})\big\|_{2}^{2} \nonumber\\
    & =  \sum_{v=1}^{d} \sum_{k=1}^{m} \sigma_{t,k}^2 (\sum_{l=1}^{m}\bp_{k,l}^2)
    \{\frac{\sum_{l=1}^{m}\bp_{k,l}[(\bw_l^{v,(t)}-\widetilde{\bw}_l^{v,(t)})-\mu_{t,k}^v] + \sum_{l=1}^{m}\bp_{k,l}\mu_{t,l}^v]}{\sigma_{t,k} \sqrt{\sum_{l=1}^{m}\bp_{k,l}^2}}\}^2 \nonumber\\
    & \leq  \sum_{v=1}^{d} \sum_{k=1}^{m} \sigma_{t,k}^2 (\sum_{l=1}^{m}\bp_{k,l}^2)
    \{\frac{\sum_{l=1}^{m}\bp_{k,l}[(\bw_l^{v,(t)}-\widetilde{\bw}_l^{v,(t)})-\mu_{t,k}^v]}{\sigma_{t,k} \sqrt{\sum_{l=1}^{m}\bp_{k,l}^2}}\}^2 
    + \mu^2 dm\cdot\sum_{k=1}^{m} \sum_{l=1}^{m}\bp_{k,l}^2\nonumber\\
    & +  \sum_{v=1}^{d} \sum_{k=1}^{m} (\sum_{l=1}^{m}\bp_{k,l}^2)
    \frac{\sum_{l_1=1}^{m}\sum_{l_2=1}^{m}\bp_{k,l_1}\bp_{k,l_2}\mu_{t,l_1}^v[(\bw_{l_1}^{v,(t)}-\widetilde{\bw}_{l_1}^{v,(t)})-\mu_{t,l_1}^v]}{ \sqrt{\sum_{l_1=1}^{m}\bp_{k,l}^2}\sqrt{\sum_{l_2=1}^{m}\bp_{k,l}^2}},
\end{align}
where $\bw_l^{v,(t)}$ and $\widetilde{\bw}_l^{v,(t)}$ stacks the $v$-th entry of the $d$-dimensional vector $\bw_l^{(t)}$ and $\widetilde{\bw}_l^{(t)}$, respectively. The last inequality holds since $\sum_{v=1}^{d}(\mu_{t,l}^v)^2\leq d\mu^2 $.

Since the weight difference is normally distributed:
\[
    \big(\bw_k^{(t)}-\widetilde{\bw}_k^{(t)}\big) \sim \mathcal{N}(\mu_{t,k}, \sigma_{t,k}^2 I_{d}),\quad k = 1, \dots, m,
\]
with $\mu_{t,k}$ satisfying $\|\mu_{t,k}\|_2^2\leq d\cdot\mu^2$ and $\sigma_{t,k}^2\in\rbb$ being bounded by $\sigma^2$,
we obtain
\[
    \frac{\sum_{l=1}^{m}\bp_{k,l}[(\bw_l^{v,(t)}-\widetilde{\bw}_l^{v,(t)})-\mu_{t,k}^v]}{\sigma_{t,k}\sqrt{\sum_{l=1}^{m}\bp_{k,l}^2}} \stackrel{ i.i.d.}{\sim} \mathcal{N}(0, 1),\quad l = 1, \dots, m.
\]
Since the sum of squared i.i.d. standard normal variables follows a Chi-Square distribution with 1 degree of freedom, we arrive at
\[
    \{\frac{\sum_{l=1}^{m}\bp_{k,l}[(\bw_l^{v,(t)}-\widetilde{\bw}_l^{v,(t)})-\mu_{t,k}^v]}{\sigma_{t,k}\sqrt{\sum_{l=1}^{m}\bp_{k,l}^2}}\}^2 {\sim} \mathcal{X}^2{(1)}, \quad l = 1, \dots, m.
\]
Furthermore, since $\forall l$ $\ebb_A[(\bw_{l}^{v,(t)}-\widetilde{\bw}_{l}^{v,(t)})-\mu_{t,l}^v]=0$, we have
\[
    \ebb_A\left[\sum_{v=1}^{d} \sum_{k=1}^{m} (\sum_{l=1}^{m}\bp_{k,l}^2)
    \frac{\sum_{l_1=1}^{m}\sum_{l_2=1}^{m}\bp_{k,l_1}\bp_{k,l_2}\mu_{t,l_1}^v[(\bw_{l_1}^{v,(t)}-\widetilde{\bw}_{l_1}^{v,(t)})-\mu_{t,l_1}^v]}{ \sqrt{\sum_{l_1=1}^{m}\bp_{k,l}^2}\sqrt{\sum_{l_2=1}^{m}\bp_{k,l}^2}} \right]=0.
\]
As a consequence,
\[
    \ebb_A\big[\big\|\bp (\bW^{(t)}-\widetilde{\bW}^{(t)})\big\|_{F}^{2}
    \leq d\sigma^2 \sum_{k=1}^{m} \lambda_k^2 + d\mu^2 \sum_{k=1}^{m} \lambda_k^2 \leq 
     d(\sigma^2+\mu^2) [1+(m-1)\lambda^2],
\]
where $1-\lambda$ is the spectral gap of the communication topology.

(2) For the second part $T_2$, we have
\begin{align}
    & T_{2} \leq \sum_{k=1}^{m} \ind_{z_{k,\zeta_t}^{(t)}\neq \tilde{z}_{k,\zeta_t}^{(t)}} 
    p\big\|\sum_{l=1}^{m}\bp_{k,l} (\bw_l^{(t)}\unaryminus\widetilde{\bw}_l^{(t)})\big\|_{2}^{2}
    \unaryplus \sum_{k=1}^{m} \ind_{z_{k,\zeta_t}^{(t)}\neq \tilde{z}_{k,\zeta_t}^{(t)}}(1+p^{-1})c_{\alpha,1}^2{\eta_t}^2 \big( f^{\frac{2\alpha}{1+\alpha}}(\bw_{k}^{(t)}; z_{k,\zeta_t}^{(t)})\unaryplus f^{\frac{2\alpha}{1+\alpha}}(\widetilde{\bw}_k^{(t)}, \tilde{z}_{k,\zeta_t}^{(t)})\big).\label[inequality]{ineq:T_2-th1}
\end{align}  
\textit{Proof.}

\cref{ineq:T_2-th1} are  mainly based on \cref{th:standard-inequal} and \cref{lem:self-bounding}:
\begin{align}
     T_{2} &= \sum_{k=1}^{m} \ind_{z_{k,\zeta_t}^{(t)}\neq \tilde{z}_{k,\zeta_t}^{(t)}}  \big[\big\|\bw_{k}^{(t+1)}-\widetilde{\bw}_{k}^{(t+1)}\big\|_{2}^{2}-\big\|\sum_{l=1}^{m}\bp_{k,l} (\bw_l^{(t)}-\widetilde{\bw}_l^{(t)})\big\|_{2}^{2}\big]\nonumber\\
     & = \sum_{k=1}^{m} \ind_{z_{k,\zeta_t}^{(t)}\neq \tilde{z}_{k,\zeta_t}^{(t)}} 
    \big[\big\|\sum_{l=1}^{m} \mathbf{P}_{k, l} \bw_{t}(l)\unaryminus\eta_{t} \nabla f\left(\bw_{k}^{(t)}; z_{k,\zeta_t}^{(t)}\right)\unaryminus\sum_{l=1}^{m} \mathbf{P}_{k, l} \widetilde{\bw}_l^{(t)}\unaryplus\eta_{t} \nabla f\big(\widetilde{\bw}_k^{(t)}, \tilde{z}_{k,\zeta_t}^{(t)}\big)\big\|_{2}^{2} \unaryminus\big\|\sum_{l=1}^{m}\bp_{k,l} (\bw_l^{(t)}\unaryminus\widetilde{\bw}_l^{(t)})\big\|_{2}^{2}\big] \nonumber\\
    & \underset{{\text {\cref{th:standard-inequal}}}}{\leq} \sum_{k=1}^{m} \ind_{z_{k,\zeta_t}^{(t)}\neq \tilde{z}_{k,\zeta_t}^{(t)}} 
    \big[p\big\|\sum_{l=1}^{m}\bp_{k,l} (\bw_l^{(t)}-\widetilde{\bw}_l^{(t)})\big\|_{2}^{2}
     + (1+p^{-1}){\eta_t}^2\big\|\nabla f\left(\bw_{k}^{(t)}; z_{k,\zeta_t}^{(t)}\right)-\nabla f\left(\widetilde{\bw}_k^{(t)}, \tilde{z}_{k,\zeta_t}^{(t)}\right)\big\|_{2}^{2}\big]\nonumber\\
    & \underset{{\text {\cref{lem:self-bounding}}}}{\leq}\underbrace{\sum_{k=1}^{m} \ind_{z_{k,\zeta_t}^{(t)}\neq \tilde{z}_{k,\zeta_t}^{(t)}} 
    \big[p\big\|\sum_{l=1}^{m}\bp_{k,l} (\bw_l^{(t)}-\widetilde{\bw}_l^{(t)})\big\|_{2}^{2}}_{ T_{2.1}}
     + (1+p^{-1})c_{\alpha,1}^2{\eta_t}^2 \big( f^{\frac{2\alpha}{1+\alpha}}(\bw_{k}^{(t)}; z_{k,\zeta_t}^{(t)})+f^{\frac{2\alpha}{1+\alpha}}(\widetilde{\bw}_k^{(t)}, \tilde{z}_{k,\zeta_t}^{(t)})\big)\big].\nonumber
\end{align}

(3)  $T_3$ can be controlled as follows:
\begin{equation}
 T_{3}\leq  \sum_{k=1}^{m} \ind_{z_{k,\zeta_t}^{(t)}= \tilde{z}_{k,\zeta_t}^{(t)}}
    \eta_t\big[ 2L\big\|\bw_{k}^{(t)}-\widetilde{\bw}_{k}^{(t)}\big\|_{2}^{2\alpha}
    + \big\|\sum_{l=1}^{m}\bp_{k,l} (\bw_l^{(t)}\unaryminus\widetilde{\bw}_l^{(t)})\big\|_2^2 \big]. \label[inequality]{ineq:T_3-th1}
\end{equation}

\textit{Proof.}

According to the H\"older continuous assumption, we have
\begin{equation}
    \sum_{k=1}^{m} \ind_{z_{k,\zeta_t}^{(t)}= \tilde{z}_{k,\zeta_t}^{(t)}} \big\|\nabla f(\bw_k^{(t)};z_{k,\zeta_t}^{(t)})-\nabla f(\widetilde{\bw}_k^{(t)};\tilde{z}_{k,\zeta_t}^{(t)})\big\|_2^2\leq L \sum_{k=1}^{m} \ind_{z_{k,\zeta_t}^{(t)} = \tilde{z}_{k,\zeta_t}^{(t)}}   \big\|\bw_{k}^{(t)}-\widetilde{\bw}_{k}^{(t)}\big\|_{2}^{2\alpha}.
\end{equation}
Consequently,
\begin{align}
    T_3 &\leq  \sum_{k=1}^{m} \ind_{z_{k,\zeta_t}^{(t)} = \tilde{z}_{k,\zeta_t}^{(t)}} \eta_t\big[L\big\|\bw_{k}^{(t)}-\widetilde{\bw}_{k}^{(t)}\big\|_{2}^{2\alpha}
    -2\big\langle\sum_{l=1}^{m}\bp_{k,l} (\bw_l^{(t)}-\widetilde{\bw}_l^{(t)}), \nabla f\left(\bw_{k}^{(t)}; z_{k,\zeta_t}^{(t)}\right)-\nabla f\left(\widetilde{\bw}_k^{(t)}, \tilde{z}_{k,\zeta_t}^{(t)}\right)\big\rangle\big] \nonumber\\
    & \leq \sum_{k=1}^{m} \ind_{z_{k,\zeta_t}^{(t)}= \tilde{z}_{k,\zeta_t}^{(t)}}
    \eta_t\big[ L\big\|\bw_{k}^{(t)}-\widetilde{\bw}_{k}^{(t)}\big\|_{2}^{2\alpha}
    +2 \big\|\sum_{l=1}^{m}\bp_{k,l} (\bw_l^{(t)}\unaryminus\widetilde{\bw}_l^{(t)})\big\|_2 \cdot \big\|\nabla f(\bw_k^{(t)};z_{k,\zeta_t}^{(t)})\unaryminus\nabla f(\widetilde{\bw}_k^{(t)};\tilde{z}_{k,\zeta_t}^{(t)})\big\|_2 \big]\nonumber\\
    &\underset{{\text {\cref{inequal:simple-inequal}}}}{\leq}
    \sum_{k=1}^{m} \ind_{z_{k,\zeta_t}^{(t)}= \tilde{z}_{k,\zeta_t}^{(t)}}
    \eta_t\big[ L\big\|\bw_{k}^{(t)}-\widetilde{\bw}_{k}^{(t)}\big\|_{2}^{2\alpha}
    + \big\|\sum_{l=1}^{m}\bp_{k,l} (\bw_l^{(t)}\unaryminus\widetilde{\bw}_l^{(t)})\big\|_2^2 +  \big\|\nabla f(\bw_k^{(t)};z_{k,\zeta_t}^{(t)})\unaryminus\nabla
    f(\widetilde{\bw}_k^{(t)};\tilde{z}_{k,\zeta_t}^{(t)})\big\|_2^2 \big]\nonumber\\
    & \leq \underbrace{\sum_{k=1}^{m} \ind_{z_{k,\zeta_t}^{(t)}= \tilde{z}_{k,\zeta_t}^{(t)}}
    \eta_t\big[ 2L\big\|\bw_{k}^{(t)}-\widetilde{\bw}_{k}^{(t)}\big\|_{2}^{2\alpha}}_{ T_{3.1}}
    + \big\|\sum_{l=1}^{m}\bp_{k,l} (\bw_l^{(t)}\unaryminus\widetilde{\bw}_l^{(t)})\big\|_2^2 \big].
\end{align}

(4) A simple combination of \cref{ineq:T_1-th1_1}, \cref{ineq:T_2-th1} and \cref{ineq:T_3-th1} provides the following:
\begin{align}
       \sum_{k=1}^{m}&\big\|\bw_{k}^{(t+1)}-\widetilde{\bw}_{k}^{(t+1)}\big\|_{2}^{2}=
       T_1+T_2+T_3\nonumber\\
        \leq &T_1  + \sum_{k=1}^{m} \ind_{z_{k,\zeta_t}^{(t)}\neq \tilde{z}_{k,\zeta_t}^{(t)}} 
        \big[p\big\|\sum_{l=1}^{m}\bp_{k,l} (\bw_l^{(t)}-\widetilde{\bw}_l^{(t)})\big\|_{2}^{2}
        + (1+p^{-1})c_{\alpha,1}^2{\eta_t}^2 \big( f^{\frac{2\alpha}{1+\alpha}}(\bw_{k}^{(t)}; z_{k,\zeta_t}^{(t)})+f^{\frac{2\alpha}{1+\alpha}}(\widetilde{\bw}_k^{(t)}, \tilde{z}_{k,\zeta_t}^{(t)})\big)\big]\nonumber\\
        & + \sum_{k=1}^{m} \ind_{z_{k,\zeta_t}^{(t)}= \tilde{z}_{k,\zeta_t}^{(t)}}
    \eta_t\big[ 2L\big\|\bw_{k}^{(t)}-\widetilde{\bw}_{k}^{(t)}\big\|_{2}^{2\alpha}
    + \big\|\sum_{l=1}^{m}\bp_{k,l} (\bw_l^{(t)}\unaryminus\widetilde{\bw}_l^{(t)})\big\|_2^2 \big].\label[inequality]{inq:T1+T2+T3}
\end{align}

Let $\scal_k$, $\scal$, $\scal^{(i)}_k$ and $\scal^{(i)}$ be constructed in \cref{def:dis-aver-stab}. We know that $\scal^{(i)}$ differs from $\scal$ by only the $i$-th element.
Consequently, at the $t$-th iterate, with a probability of $1-\frac{1}{N}$, the example $z_{k,\zeta_t}^{(t)}$ selected by D-SGD on worker $k$ in both
$\scal_k$ and $\scal_k^{(i)}$ is the same (i.e., $z_{k,\zeta_t}^{(t)} = \tilde{z}_{k,\zeta_t}^{(t)}$); and with a probability of $\frac{1}{N}$, the selected example is different (i.e., $z_{k,\zeta_t}^{(t)} \neq \tilde{z}_{k,\zeta_t}^{(t)}$)\footnote{Note that the selected example is different only when $z_{k,\zeta_t}^{(t)}=z_i$ and $\tilde{z}_{k,\zeta_t}^{(t)}=\tilde{z}_i$, where $z_i=\scal\setminus\scal^{(i)}$ and $\tilde{z}_i=\scal^{(i)}\setminus\scal$.}. 

Since $A$ is independent of $k$, $\ebb_A (T_{2.1})$ and $\ebb_A (T_{3.1})$ can be controlled accordingly as follows:
\begin{align}
    \ebb_A (T_{2.1}) &= \ebb_A\big[p\sum_{k=1}^{m} \ind_{z_{k,\zeta_t}^{(t)}\neq \tilde{z}_{k,\zeta_t}^{(t)}} 
    \big\|\sum_{l=1}^{m}\bp_{k,l} (\bw_l^{(t)}-\widetilde{\bw}_l^{(t)})\big\|_{2}^{2}\big]\nonumber\\
    & = \frac{p}{N}\sum_{k=1}^{m}\ebb_A \big[\big\|\sum_{l=1}^{m}\bp_{k,l}(\bw_l^{(t)}-\widetilde{\bw}_l^{(t)})\big\|_{2}^{2}\big]
    \underset{{\text {\cref{ineq:T_1-th1_1}}}}{\leq} \frac{p}{N}d(\sigma^2 + \mu^2)\big[(m-1)\lambda^2+1\big]\label{inq:T_2.1}
\end{align}
where the proof of the last inequality is analogous to part (1).

By the concavity of the mapping $x\mapsto x^{\alpha}\ (\alpha\in[0,1])$, we have
\begin{align}
    \ebb_A (T_{3.1}) 
    &\leq 2\eta_t L (1-\frac{1}{N})\ebb_A\big[\sum_{k=1}^{m} \big\|\bw_{k}^{(t)}-\widetilde{\bw}_{k}^{(t)}\big\|_{2}^{2\alpha}\big]\nonumber\\ 
    &\leq 
    2\eta_t L (1-\frac{1}{N})\{\ebb_A\big[ \sum_{k=1}^{m} \big\|\bw_{k}^{(t)}-\widetilde{\bw}_{k}^{(t)}\big\|_{2}^{2}\big]\}^{\alpha} 
    \leq 2\eta_t L (1-\frac{1}{N})\ebb_A\big[ \sum_{k=1}^{m} \big\|\bw_{k}^{(t)}-\widetilde{\bw}_{k}^{(t)}\big\|_{2}^{2}\big].
    \label{inq:T_3.2}
\end{align}
The last inequality holds if $m\geq \frac{1}{d\mu_0^2}$, where $d\mu_0^2$ is the lower bound of $ \|\mu_{t,k}\|_2^2\ (k=1\dots m)$, which leads to $\ebb_A\big[ \sum_{k=1}^{m} \big\|\bw_{k}^{(t)}-\widetilde{\bw}_{k}^{(t)}\big\|_{2}^{2}\big]\geq 1$. The condition $m\geq \frac{1}{d\mu_0^2}$ can be easily satisfied in training overparameterized models in a decentralized manner, since both $m$ and $d$ are large in these cases.

We can further replace $z_{k,\zeta_t}^{(t)}$ with $z_i$ and replace $\tilde{z}_{k,\zeta_t}^{(t)}$ with $\tilde{z}_i$ in the case when the selected example is different (i.e., $z_{k,\zeta_t}^{(t)} \neq \tilde{z}_{k,\zeta_t}^{(t)}$). Then, taking the expectation on both sides of \cref{inq:T1+T2+T3} provides
\begin{align}
       \ebb_A&\big[\sum_{k=1}^{m}\big\|\bw_{k}^{(t+1)}\unaryminus\widetilde{\bw}_{k}^{(t+1)}\big\|_{2}^{2}\big]
       \leq 2\eta_t L (1-\frac{1}{N})\ebb_A\big[ \sum_{k=1}^{m} \big\|\bw_{k}^{(t)}-\widetilde{\bw}_{k}^{(t)}\big\|_{2}^{2}\big]\nonumber\\
        & \unaryplus [1 \unaryplus \frac{p}{N}\unaryplus(1\unaryminus\frac{1}{N})\eta_t]d(\sigma^2 \unaryplus \mu^2)\big[(m\unaryminus 1)\lambda^2+1\big]  
        + \frac{1}{N} \sum_{k=1}^{m}
        \big[(1\unaryplus p^{-1})c_{\alpha,1}^2{\eta_t}^2 \big( f^{\frac{2\alpha}{1\unaryplus\alpha}}(\bw_{k}^{(t)}; z_i)\unaryplus f^{\frac{2\alpha}{1+\alpha}}(\widetilde{\bw}_k^{(t)}, \tilde{z}_i)\big)\big].
\end{align}
Knowing that $z_i$ and $\tilde{z}_i$ follow the same distribution, we have
  \[
  \ebb_{\scal,\scal^{(i)},A}\big[f^{\frac{2\alpha}{1+\alpha}}(\widetilde{\bw}_k^{(t)};\tilde{z}_i)\big]=\ebb_{\scal,A}\big[f^{\frac{2\alpha}{1+\alpha}}(\bw_k^{(t)};z_i)\big].
  \]
Note that $\{\eta_t\}$ is an non-increasing sequence. As a consequence, 
\begin{align}
       \sum_{k=1}^{m}\ebb_{\scal,\scal^{(i)},A}&\big[\big\|\bw_{k}^{(t+1)}-\widetilde{\bw}_{k}^{(t+1)}\big\|_{2}^{2}\big]
       \leq  \underbrace{2\eta_0 L (1-\frac{1}{N})}_{\triangleq C \text { (scaling coefficient)}}\ebb_{\scal,\scal^{(i)},A}\big[\sum_{k=1}^{m}\big\|\bw_{k}^{(t)}-\widetilde{\bw}_{k}^{(t)}\big\|_{2}^{2}\big]  \nonumber\\
        &+ \underbrace{[1 \unaryplus \frac{p}{N}\unaryplus(1\unaryminus\frac{1}{N})\eta_t]d(\sigma^2 \unaryplus \mu^2)\big[(m\unaryminus1)\lambda^2+1\big]}_{\text {topology-dependent}}
        + \underbrace{\frac{2}{N} 
        \big[(1\unaryplus p^{-1})c_{\alpha,1}^2{\eta_t}^2 \sum_{k=1}^{m}\ebb_{\scal,A}\big[ f^{\frac{2\alpha}{1+\alpha}}(\bw_{k}^{(t)}; z_i)\big]\big]}_{\text {topology-independent}}.\label[inequality]{inq:T1+T2+T3_2}
\end{align}
Multiplying both sides of \cref{inq:T1+T2+T3_2} with ${C}^{-(t+1)}$ provides
\begin{align}
       {C}^{-(t+1)}&\sum_{k=1}^{m}\ebb_{\scal,\scal^{(i)},A}\big[\big\|\bw_{k}^{(t+1)}-\widetilde{\bw}_{k}^{(t+1)}\big\|_{2}^{2}\big]
       \leq  {C}^{-t}\ebb_{\scal,\scal^{(i)},A}\big[\sum_{k=1}^{m}\big\|\bw_{k}^{(t)}-\widetilde{\bw}_{k}^{(t)}\big\|_{2}^{2}\big]  \nonumber\\
        &+ {C}^{-(t+1)} \big\{ [1 \unaryplus \frac{p}{N}\unaryplus(1-\frac{1}{N})\eta_t]d(\sigma^2 \unaryplus \mu^2)\big[(m\unaryminus1)\lambda^2\unaryplus1\big]
        \unaryplus \frac{2}{N} 
        \big[(1\unaryplus p^{-1})c_{\alpha,1}^2{\eta_t}^2 \sum_{k=1}^{m}\ebb_{\scal,A}\big[ f^{\frac{2\alpha}{1\unaryplus\alpha}}(\bw_{k}^{(t)}; z_i)\big]\big]\big\}.
\end{align}
Taking the summation over the iteration $\tau$, we can write
\begin{align}
       &\sum_{\tau=0}^{t}{C}^{-(\tau+1)}\sum_{k=1}^{m}\ebb_{\scal,\scal^{(i)},A}\big[\big\|\bw_{k}^{(\tau+1)}-\widetilde{\bw}_{k}^{(\tau+1)}\big\|_{2}^{2}\big]
       \leq  \sum_{\tau=0}^{t}{C}^{-\tau}\ebb_{\scal,\scal^{(i)},A}\big[\sum_{k=1}^{m}\big\|\bw_{k}^{(\tau)}-\widetilde{\bw}_{k}^{(\tau)}\big\|_{2}^{2}\big]  \nonumber\\
        &+ \sum_{\tau=0}^{t}{C}^{-(\tau+1)} \big\{ [1 \unaryplus \frac{p}{N}\unaryplus(1\unaryminus\frac{1}{N})\eta_\tau]d(\sigma^2 \unaryplus \mu^2)\big[(m\unaryminus 1)\lambda^2\unaryplus 1\big]
        + \frac{2}{N} 
        \big[(1\unaryplus p^{-1})c_{\alpha,1}^2{\eta_\tau}^2 \sum_{k=1}^{m}\ebb_{\scal,A}\big[ f^{\frac{2\alpha}{1+\alpha}}(\bw_{k}^{(\tau)}; z_i)\big]\big]\big\}.\label{inq:stab-wo-sum}
\end{align}
Since for all $ k, \bw_{1}(k)=\widetilde{\bw}_{1}(k)=0$ (see Definition \cpageref{def:dec-sgd}), we have
\begin{align}
       &\sum_{k=1}^{m}\ebb_{\scal,\scal^{(i)},A}\big[\big\|\bw_{k}^{(t+1)}-\widetilde{\bw}_{k}^{(t+1)}\big\|_{2}^{2}\big]\nonumber\\
       &\leq \sum_{\tau=0}^{t} {C}^{t-\tau} \big\{ [1 \unaryplus \frac{p}{N}\unaryplus(1\unaryminus\frac{1}{N})\eta_\tau]d(\sigma^2 \unaryplus \mu^2)\big[(m\unaryminus 1)\lambda^2\unaryplus 1\big]
        + \frac{2}{N} 
        \big[(1\unaryplus p^{-1})c_{\alpha,1}^2{\eta_\tau}^2 \sum_{k=1}^{m}\ebb_{\scal,A}\big[ f^{\frac{2\alpha}{1+\alpha}}(\bw_{k}^{(\tau)}; z_i)\big]\big]\big\}.
\end{align}
Since the mapping $x\mapsto x^{\frac{2\alpha}{1+\alpha}}$ is concave, we have $\frac{1}{N}\sum_{i=1}^{N} f^{\frac{2\alpha}{1+\alpha}}(\bw_{k}^{(\tau)}, z_i)\leq F_{\scal}^{\frac{2\alpha}{1+\alpha}}(\bw_k^{(\tau)})$. Then the distributed on-average stability of D-SGD can be controlled as follows:
\begin{align}
       \frac{1}{mN}&\sum_{i=1}^{N}\sum_{k=1}^{m}\ebb_{\scal,\scal^{(i)},A}\big[\big\|\bw_{k}^{(t+1)}-\widetilde{\bw}_{k}^{(t+1)}\big\|_{2}^{2}\big]\nonumber\\
       &\leq \sum_{\tau=0}^{t} {C}^{t-\tau} \big\{ [1 \unaryplus \frac{p}{N}\unaryplus(1\unaryminus\frac{1}{N})\eta_\tau]d(\sigma^2 \unaryplus \mu^2)\big[(1\unaryminus \frac{1}{m})\lambda^2\unaryplus \frac{1}{m}\big]
        + \frac{2}{N} 
        \big[(1\unaryplus p^{-1})c_{\alpha,1}^2{\eta_\tau}^2 \frac{1}{m}\sum_{k=1}^{m}\ebb_{\scal,A}\big[ F_{\scal}^{\frac{2\alpha}{1+\alpha}}(\bw_k^{(\tau)})\big]\big]\big\}.\label[inequality]{inq:on-average-holder}
\end{align}
The proof is complete.

{\color{magenta}\qed}

\textit{\textbf{Proof of \cref{th:stab-fix-eta}.}}

With constant step size $\eta_t\equiv\eta \leq \frac{1}{2L} (1-\frac{2}{m})$, $\sum_{\tau=0}^{t} {C}^{t-\tau}$ can be written as
\[
    \sum_{\tau=0}^{t} {C}^{t-\tau} = \sum_{\tau=0}^{t} {[2\eta L (1-\frac{1}{N})]}^{t-\tau} = \sum_{\tau=0}^{t-1} {[2\eta L (1-\frac{1}{N})]}^{\tau} = \frac{1- [2\eta L (1-\frac{1}{N})]^t}{1- 2\eta L (1-\frac{1}{N})}.
\]

Consequently, the distributed on-average stability of D-SGD is bounded as follows:
\begin{flalign}
        \frac{1}{mn}\sum_{i=1}^{n}\sum_{k=1}^{m}\ebb_{\scal,\scal^{(i)},A}&\big[\big\|\bw_{k}^{(t+1)}-\widetilde{\bw}_{k}^{(t+1)}\big\|_{2}^{2}\big]\nonumber\\
        & \leq \frac{1- [2\eta L (1-\frac{1}{N})]^t}{1- 2\eta L (1-\frac{1}{N})} \big\{ [1 \unaryplus \frac{p}{N}\unaryplus(1\unaryminus\frac{1}{N})\eta]d(\sigma^2 \unaryplus \mu^2)\big[(1\unaryminus \frac{1}{m})\lambda^2\unaryplus \frac{1}{m}\big]
        + \frac{2}{N} 
        \big[(1\unaryplus p^{-1})c_{\alpha,1}^2{\eta}^2 \epsilon_{\scal}]\big\},
\end{flalign}
where $\epsilon_{\scal}$ denotes the upper bound of $ \frac{1}{m}\sum_{k=1}^{m}\ebb_{\scal,A}\big[F_{\scal}^{\frac{2\alpha}{1+\alpha}}(\bw_{k}^{(t)})\big]\ \forall t$.

Letting $t\rightarrow \infty$ further provides
\begin{flalign}
        \frac{1}{mn}\sum_{i=1}^{n}\sum_{k=1}^{m}\ebb_{\scal,\scal^{(i)},A}&\big[\big\|\bw_{k}^{(t+1)}-\widetilde{\bw}_{k}^{(t+1)}\big\|_{2}^{2}\big]\nonumber\\
        & \leq \frac{1}{1-2\eta L (1-\frac{1}{N})} \big\{ [1 \unaryplus \frac{p}{N}\unaryplus(1\unaryminus\frac{1}{N})\eta_\tau]d(\sigma^2 \unaryplus \mu^2)\big[(1\unaryminus \frac{1}{m})\lambda^2\unaryplus \frac{1}{m}\big]
        + \frac{2}{N} 
        \big[(1\unaryplus p^{-1})c_{\alpha,1}^2{\eta_\tau}^2 \epsilon_{\scal}]\big\}.
\end{flalign}

{\color{magenta}\qed}

\subsection{Generalization of D-SGD}\label{sec:pf-gen-stab}

\textit{\textbf{Proof of \cref{prop:gen-stab}.}}

We denote $A(\scal)$ as the model produced by algorithm $A$ based on the training dataset $\scal$.

 To begin with, we can write
  \begin{align}
    \ebb_{\scal,A}\big[F(A(\scal))-F_{\scal}(A(\scal))\big] & = \frac{1}{N}\sum_{i=1}^{N}\ebb_{\scal,\scal^{(i)},A}\big[F(A(\scal^{(i)}))-F_{\scal}(A(\scal))\big] \notag\\
    & = \frac{1}{N}\sum_{i=1}^{N}\ebb_{\scal,\scal^{(i)},A}\big[\ebb_{\tilde{z}\sim D}(f(A(\scal^{(i)});\tilde{z}_i))-f(A(\scal);z_i)\big]\\
     & = \frac{1}{N}\sum_{i=1}^{N}\ebb_{\scal,\scal^{(i)},A}\big[f(A(\scal^{(i)});z_i)-f(A(\scal);z_i)\big],\label{gen-model-stab-1}
  \end{align}
  where the first line follows from noticing that $\ebb_{\scal,A}\big[F(A(\scal))\big]=\ebb_{\scal^{(i)},A}\big[F(A(\scal^{(i)}))\big]$ and the last identity holds since $A(\scal^{(i)})$ is independent of $z_i$ and thus $\ebb_{\scal,\scal^{(i)},A}\big[\ebb_{\tilde{z}\sim \scal^{(i)}}(f(A(\scal^{(i)});\tilde{z})\big] = \ebb_{\scal,\scal^{(i)},A}\big[f(A(\scal^{(i)});z_i)\big]$.

\cref{lemma: upper-bounding-f(x)-f(y)} and the concavity of the $x\mapsto x^{\frac{1+\alpha}{2}}$ further guarantee
  \begin{align*}
    \ebb_{\scal,A}\big[F(A(\scal))-F_{\scal}(A(\scal))\big]  \leq &\frac{1}{N}\sum_{i=1}^{N}\ebb_{\scal,\scal^{(i)},A}\big[\frac{L}{1+\alpha}\|A(\scal)-A(\scal^{(i)})\|_2^{1+\alpha}\big] \\
    & + \frac{1}{2}\frac{1}{N}\sum_{i=1}^{N}\ebb_{\scal,\scal^{(i)},A}\big[\|A(\scal)-A(\scal^{(i)})\|^2_2\big] + \frac{1}{2}\ebb_{\scal,A}\big[\frac{1}{N}\sum_{i=1}^{N}\|\nabla f(A(\scal);z_i)\|^2_2\big]\\
     \leq &\frac{L}{1+\alpha}\{\frac{1}{N}\sum_{i=1}^{N}\ebb_{\scal,\scal^{(i)},A}\big[\|A(\scal)-A(\scal^{(i)})\|^2_2\big]\}^{\frac{1+\alpha}{2}}\\
    & + \frac{1}{2}\frac{1}{N}\sum_{i=1}^{N}\ebb_{\scal,\scal^{(i)},A}\big[\|A(\scal)-A(\scal^{(i)})\|^2_2\big] + \frac{1}{2}\ebb_{\scal,A}\big[\frac{1}{N}\sum_{i=1}^{N}\|\nabla f(A(\scal);z_i)\|^2_2\big].\label{inq:refined-lei}
  \end{align*}
  
Finally, consider $\frac{1}{m}\sum_{k=1}^m\bw_k^{(t)}$ as an output of algorithm $A$ on dataset $\scal$. For the sake of simplicity, we expect $\forall t$ $\ebb_{\scal,\scal^{(i)},A}\|\frac{1}{m}\sum_{k=1}^m\bw_k^{(t)}-\frac{1}{m}\sum_{k=1}^m\widetilde{\bw}_{k}^{(t)}\|_2\leq 1$ without loss of generality. It is a mild assumption since $\scal$ and $\scal^{(i)}$ differ by only one data point. We can then complete the proposition by the convexity of vector 2-norm and square function:
    \begin{align}
        \ebb_{\scal,A}&\big[F(\frac{1}{m}\sum_{k=1}^m\bw_k^{(t)})-F_\scal(\frac{1}{m}\sum_{k=1}^m\widetilde{\bw}_{k}^{(t)})\big]\nonumber\\
        &\leq (\frac{L}{1+\alpha}\unaryplus\frac{1}{2})\{\frac{1}{N}\sum_{i=1}^{n}\ebb_{\scal,\scal^{(i)},A}\big[\|\frac{1}{m}\sum_{k=1}^m\bw_k^{(t)}\unaryminus\frac{1}{m}\sum_{k=1}^m\widetilde{\bw}_{k}^{(t)}\|_2^2\big]\}^{\frac{1+\alpha}{2}}
        \unaryplus\frac{1}{2}\ebb_{\scal,A}\big[\frac{1}{N}\sum_{i=1}^{N}\|\nabla f(\frac{1}{m}\sum_{k=1}^m\bw_k^{(t)};z_i)\|^2_2\big]\nonumber\\
        &\leq (\frac{L}{1+\alpha}\unaryplus\frac{1}{2})\{\frac{1}{mN}\sum_{i=1}^{N}\sum_{k=1}^{m}\ebb_{\scal,\scal^{(i)},A}\big[\|\bw_k^{(t)}-\widetilde{\bw}_{k}^{(t)}\|_2^2\big]\big\}^{\frac{1+\alpha}{2}}
        \unaryplus\underbrace{\frac{1}{2}\ebb_{\scal,A}\big[\frac{1}{N}\sum_{i=1}^{N}\|\nabla f(\frac{1}{m}\sum_{k=1}^m\bw_k^{(t)};z_i)\|^2_2\big]}_{\text {empirical gradient norm}}.
    \end{align}

{\color{magenta}\qed}

\textit{\textbf{Proof of \cref{thm:gen-expected}.}}

We start by rewriting \cref{inq:on-average-holder} as
\begin{align}
        \frac{1}{mn}&\sum_{i=1}^{n}\sum_{k=1}^{m}\ebb_{\scal,\scal^{(i)},A}\big[\big\|\bw_{k}^{(t+1)}-\widetilde{\bw}_{k}^{(t+1)}\big\|_{2}^{2}\big]\nonumber\\
        & \leq \sum_{\tau=0}^{t} {C}^{t-\tau} \big\{ [1 \unaryplus \frac{p}{n}\unaryplus(1\unaryminus\frac{1}{n})\eta_\tau]d(\sigma^2 \unaryplus \mu^2)\big[(1\unaryminus \frac{1}{m})\lambda^2\unaryplus \frac{1}{m}\big]
        + \frac{1}{n} 
        \big[2(1\unaryplus p^{-1})c_{\alpha,1}^2{\eta_\tau}^2 \frac{1}{m}\sum_{k=1}^{m}\ebb_{\scal,A}\big[ F_{\scal}^{\frac{2\alpha}{1+\alpha}}(\bw_k^{(\tau)})\big]\big]\big\}.\label[inequality]{inq:pf-gen}
\end{align}
To facilitate subsequent analysis, we denote $T_{\text {dec}} = \sum_{\tau=0}^{t} {C}^{t-\tau}[1 \unaryplus \frac{p}{n}\unaryplus(1\unaryminus\frac{1}{n})\eta_\tau]d(\sigma^2 \unaryplus \mu^2)\big[(1\unaryminus \frac{1}{m})\lambda^2\unaryplus \frac{1}{m}\big]$ and $T_{\text {avg}} = \sum_{\tau=0}^{t} {C}^{t-\tau}\big[2(1\unaryplus p^{-1})c_{\alpha,1}^2{\eta_\tau}^2 \frac{1}{m}\sum_{k=1}^{m}\ebb_{\scal,A}\big[ F_{\scal}^{\frac{2\alpha}{1+\alpha}}(\bw_k^{(\tau)})\big]\big]$.

A combination of \cref{inq:on-average-holder} and \cref{prop:gen-stab} yields
\begin{align}
        \ebb_{\scal,A}&\big[F(\frac{1}{m}\sum_{k=1}^m\bw_{k}^{(t+1)})-F_\scal(\frac{1}{m}\sum_{k=1}^m\bw_{k}^{(t+1)})\big]\nonumber\\
        &\leq (\frac{L}{1+\alpha}\unaryplus\frac{1}{2})\big(\frac{1}{n}T_{\text {avg}} + T_{\text {dec}})^{\frac{1+\alpha}{2}}
        + \frac{1}{2N}\sum_{i=1}^{N}\ebb_{\scal,A}\big[\|\nabla f(\frac{1}{m}\sum_{k=1}^m\bw_k^{(t)};z_i)\|^2_2\big].
\end{align}

Without loss of generality, we assume that the local training sample size satisfies $n\geq \frac{T_{\text {avg}}}{T_{\text {dec}}}$. Then the first term in the right hand side of \cref{inq:pf-gen} can be bounded as
\begin{align}
        (\frac{L}{1+\alpha}\unaryplus\frac{1}{2})(\frac{1}{n}T_{\text {avg}} + T_{\text {dec}})^{\frac{1+\alpha}{2}}
        & =  (\frac{L}{1+\alpha}\unaryplus\frac{1}{2})(\frac{T_{\text {avg}}}{n})^{\frac{1+\alpha}{2}}\big(1 + \frac{T_{\text {dec}} n}{T_{\text {avg}}})^{\frac{1+\alpha}{2}}\nonumber\\
        & \leq (\frac{L}{1+\alpha}\unaryplus\frac{1}{2})(\frac{T  _{\text {avg}}}{n})^{\frac{1+\alpha}{2}}[1+(\frac{T_{\text {dec}} n}{T_{\text {avg}}})^{\frac{1+\alpha}{2}}]\nonumber\\
        &  = (\frac{L}{1+\alpha}\unaryplus\frac{1}{2})[(\frac{T_{\text {avg}}}{n})^{\frac{1+\alpha}{2}}+  T_{\text {dec}}^\frac{1+\alpha}{2}],
\end{align}
Since the inequality $(1+x)^{\frac{1+\alpha}{2}}\leq 2^{\frac{1+\alpha}{2}} - 1 + x^{\frac{1+\alpha}{2}} \leq 1 + x^{\frac{1+\alpha}{2}}$ holds for all $x\geq 1$ and $\alpha\in[0,1]$.

Consequently, the generalization bound of D-SGD can be controlled as
\begin{align}
        \ebb_{\scal,A}\big[F&(\frac{1}{m}\sum_{k=1}^m\bw_{k}^{(t+1)})-F_\scal(\frac{1}{m}\sum_{k=1}^m\bw_{k}^{(t+1)})\big]\nonumber\\
        \leq & (\frac{L}{1+\alpha}\unaryplus\frac{1}{2})[(\frac{T_{\text {avg}}}{n})^{\frac{1+\alpha}{2}}+  T_{\text {dec}}^\frac{1+\alpha}{2}]
        + \frac{1}{2N}\sum_{i=1}^{N}\ebb_{\scal,A}\big[\|\nabla f(\frac{1}{m}\sum_{k=1}^m\bw_k^{(t)};z_i)\|^2_2\big]\nonumber\\
         \leq & \frac{1}{2N}\sum_{i=1}^{N}\ebb_{\scal,A}\big[\|\nabla f(\frac{1}{m}\sum_{k=1}^m\bw_k^{(t)};z_i)\|^2_2\big]
        + (\frac{L}{1+\alpha}\unaryplus\frac{1}{2}) {\{\frac{1}{n}\sum_{\tau=0}^{t} {C}^{t-\tau}\big[2(1+ p^{-1})c_{\alpha,1}^2{\eta_\tau}^2 \epsilon_{\scal}\big]\}}^{\frac{1+\alpha}{2}}\nonumber\\
        &+ (\frac{L}{1+\alpha}\unaryplus\frac{1}{2}) {\{\sum_{\tau=0}^{t} {C}^{t-\tau}[1 + \frac{p}{n}+(1-\frac{1}{n})\eta_\tau]d(\sigma^2 + \mu^2)\big[(1- \frac{1}{m})\lambda^2+ \frac{1}{m}\big]\}}^{\frac{1+\alpha}{2}}.\label[inequality]{gen-dsgd-1}
\end{align}
where  $\epsilon_{\scal}$ denotes the upper bound of $ \frac{1}{m}\sum_{k=1}^{m}\ebb_{\scal,A}\big[F_{\scal}^{\frac{2\alpha}{1+\alpha}}(\bw_{k}^{(t)})\big]\ \forall t$.

If we set the $\eta_t\equiv\eta \leq \frac{1}{2L}$, \cref{gen-dsgd-1} can be further written as
\begin{align}
        \ebb_{\scal,A}\big[F&(\frac{1}{m}\sum_{k=1}^m\bw_{k}^{(t+1)})-F_\scal(\frac{1}{m}\sum_{k=1}^m\bw_{k}^{(t+1)})\big] \nonumber\\
         \leq& \frac{\frac{L}{1+\alpha}\unaryplus\frac{1}{2}}{{[1\unaryminus2\eta L (1\unaryminus\frac{1}{n})]}^{\frac{1+\alpha}{2}}} 
        \{\mathcal{O}{((\frac{{\epsilon_{\scal}}}{n})^{\frac{1+\alpha}{2}} )}
        + \mathcal{O}{({[(1\unaryminus \frac{1}{m})\lambda^2\unaryplus \frac{1}{m}]}^{\frac{1+\alpha}{2}})}\}
        \frac{1}{2N}\sum_{i=1}^{N}\ebb_{\scal,A}\big[\|\nabla f(\frac{1}{m}\sum_{k=1}^m\bw_k^{(t)};z_i)\|^2_2\big]\nonumber\\
         \leq &\frac{\frac{L}{1+\alpha}\unaryplus\frac{1}{2}}{{[1\unaryminus2\eta L (1\unaryminus\frac{1}{n})]}^{\frac{1+\alpha}{2}}} 
        \{\mathcal{O}{((\frac{{\epsilon_{\scal}}}{n})^{\frac{1+\alpha}{2}} )}
        + \mathcal{O}{({ \lambda^{1+\alpha}\unaryplus m^{-\frac{1+\alpha}{2}}})}\}
        +\frac{1}{2N}\sum_{i=1}^{N}\ebb_{\scal,A}\big[\|\nabla f(\frac{1}{m}\sum_{k=1}^m\bw_k^{(t)};z_i)\|^2_2\big],
\end{align}
which completes the proof.

{\color{magenta}\qed}

 \subsection{Implications}\label{sec:pf-imp}
 
\textit{\textbf{ Proof of \cref{thm:consensus-control}}}.

\cref{inq:on-average-holder} shows that in the smooth settings ($\alpha=1$), the distributed on-average stability of D-SGD is bounded as
\begin{align}
       \frac{1}{mn}&\sum_{i=1}^{n}\sum_{k=1}^{m}\ebb_{\scal,\scal^{(i)},A}\big[\big\|\bw_{k}^{(t+1)}-\widetilde{\bw}_{k}^{(t+1)}\big\|_{2}^{2}\big]\nonumber\\
       &\leq \sum_{\tau=0}^{t} {C}^{t-\tau} \big\{ [1 \unaryplus \frac{p}{N}\unaryplus(1\unaryminus\frac{1}{N})\eta_\tau]d(\sigma^2 \unaryplus \mu^2)\big[(1\unaryminus \frac{1}{m})\lambda_\tau^2\unaryplus \frac{1}{m}\big]
        + \frac{2}{N} 
        \big[(1\unaryplus p^{-1})c_{1,1}^2{\eta_\tau}^2 \frac{1}{m}\sum_{k=1}^{m}\ebb_{\scal,A}\big[ F_{\scal}(\bw_k^{(\tau)})\big]\big]\big\},\label[inequality]{inq:stab-smooth}
\end{align}
where $C=2\eta L (1-\frac{1}{n})$.

\textbf{Our goal} is to prove that the upper bound of the stability increase with the number of iterations that we start to control the ``consensus distance''.

According to the descent lemma in \citet{pmlr-v119-koloskova20a}, the empirical risk of the consensus model can be bounded by the consensus distance as follows:

\begin{equation}
\ebb_{A}f(\overline{\bw}^{(\tau+1)};z_i) \leq  \ebb_{A}f(\overline{\bw}^{(\tau)};z_i)+\eta L^{2}
\underbrace{\frac{1}{m}\sum_{k=1}^{m}\|\bw_{k}^{(\tau)}-{\overline{\bw}^{(\tau)}}\|_{2}^{2}}_{\text {consensus distance}}
+\ebb_{A}\frac{L}{n} \eta^{2} \underbrace{\|\frac{1}{m} \sum_{k=1}^{m} \nabla f(\bw_{k}^{(\tau)}; z_i)\|_{2}^{2}}_{\text {norm of the averaged gradient}},
\end{equation}
where $\overline{\bw}^{(\tau)}=\frac{1}{m}\sum_{k=1}^m\bw_k^{(\tau)}$.

Due to the fact that the gradient of $f$ w.r.t. the first parameter is bounded by $B$ and the square of the vector 2-norm $\|\cdot\|^2_2$ is convex, we have
\begin{equation}
    \frac{1}{m}\sum_{k=1}^{m}\ebb_{\scal,A}\big[ F_{\scal}(\overline{\bw}^{(\tau)})\big]
    \leq \eta L^{2}\sum_{\nu=0}^{\tau}\frac{1}{m}
\sum_{k=1}^{m}\|{\overline{\bw}^{(\nu)}}-\bw_{k}^{(\nu)}\|_{2}^{2}
+\frac{L}{m} \eta^{2} \tau B^2.\label[inequality]{descent-lemma}
\end{equation}

To connect the stability upper bound in \cref{inq:stab-smooth}, we perform the Taylor expansion of $\frac{1}{m}\sum_{k=1}^{m}\ebb_{\scal,A}\big[ F_{\scal}(\cdot)\big]$ around $\overline{\bw}^{(\tau)}$:
\begin{align}
    \frac{1}{m}\sum_{k=1}^{m}\ebb_{\scal,A}\big[ F_{\scal}(\bw_k^{(\tau)})\big] 
    = &\frac{1}{m}\sum_{k=1}^{m}\ebb_{\scal,A}\big[ F_{\scal}(\overline{\bw}^{(\tau)})\big] + \frac{1}{m}\sum_{k=1}^{m}\big[\ebb_{\scal,A}\big[\nabla F_{\scal}(\overline{\bw}^{(\tau)})\big]\big]\tran(\bw_k^{(\tau)}-\overline{\bw}^{(\tau)}) \nonumber\\
    &  +
    (\bw_k^{(\tau)}-\overline{\bw}^{(\tau)})\tran\frac{1}{m}\sum_{k=1}^{m}\ebb_{\scal,A}\big[\nabla^2 F_{\scal}(\overline{\bw}^{(\tau)})\big] (\bw_k^{(\tau)}-\overline{\bw}^{(\tau)})
    +\mathcal{O}{(\| \bw_k^{(\tau)}-\overline{\bw}^{(\tau)}\|^3_2)}.
\end{align}

According to \cref{ass:lipschitz}, the gradient of $F_{\scal}$ w.r.t. the first parameter is bounded by $B$. Consequently, the averaged empirical loss $\frac{1}{m}\sum_{k=1}^{m}\ebb_{\scal,A}\big[ F_{\scal}(\bw_k^{(\tau)})\big]$ can be bounded as
\begin{align}
    \frac{1}{m}&\sum_{k=1}^{m}\ebb_{\scal,A}\big[ F_{\scal}(\bw_k^{(\tau)})\big]\leq \frac{1}{m}\sum_{k=1}^{m}\ebb_{\scal,A}\big[ F_{\scal}(\overline{\bw}^{(\tau)})\big] + \|\underbrace{\frac{1}{m}\sum_{k=1}^{m}(\bw_k^{(\tau)}-\overline{\bw}^{(\tau)})}_{=0}\|_2\underbrace{\|\ebb_{\scal,A}\bigg[\nabla F_{\scal}(\overline{\bw}^{(\tau)})\bigg]\|_2}_{\leq B} \nonumber\\
    &  + (\bw_k^{(\tau)}-\overline{\bw}^{(\tau)})\tran\frac{1}{m}\sum_{k=1}^{m}\ebb_{\scal,A}\big[\nabla^2 F_{\scal}(\overline{\bw}^{(\tau)})\big](\bw_k^{(\tau)}-\overline{\bw}^{(\tau)})
    +\mathcal{O}{(\| \bw_k^{(\tau)}-\overline{\bw}^{(\tau)}\|^3_2)}\nonumber\\
    & \leq \frac{1}{m}\sum_{k=1}^{m}\ebb_{\scal,A}\big[ F_{\scal}(\overline{\bw}^{(\tau)})\big] + L\underbrace{\frac{1}{m}\sum_{k=1}^{m}\|\bw_{k}^{(\tau)}-{\overline{\bw}^{(\tau)}}\|_{2}^{2}}_{\text {consensus distance}} + \mathcal{O}{(\| \bw_k^{(\tau)}-\overline{\bw}^{(\tau)}\|^3_2)}.\label[inequality]{emp-and-consensus}
\end{align}
The last inequality holds since the smooth condition
\begin{equation}
\|\nabla f(x)-\nabla f(y)\|_2 \leq L\|x-y\|_2  \Longleftrightarrow \nabla^{2} f \preceq L I,
\end{equation}
and thus we have
\begin{align}
     (\bw_k^{(\tau)}-\overline{\bw}^{(\tau)})\tran&\frac{1}{m}\sum_{k=1}^{m}\ebb_{\scal,A}\big[\nabla^2 F_{\scal}(\overline{\bw}^{(\tau)})\big](\bw_k^{(\tau)}-\overline{\bw}^{(\tau)})\nonumber\\
    &\leq L \frac{1}{m}\sum_{k=1}^{m}(\bw_k^{(\tau)}-\overline{\bw}^{(\tau)})\tran I(\bw_k^{(\tau)}-\overline{\bw}^{(\tau)})
    = L\frac{1}{m}\sum_{k=1}^{m}\|\bw_{k}^{(\tau)}-{\overline{\bw}^{(\tau)}}\|_{2}^{2}.
\end{align}If we omit the third-order difference, a combination of \cref{descent-lemma} and \cref{emp-and-consensus} provides
\begin{equation}
    \frac{1}{m}\sum_{k=1}^{m}\ebb_{\scal,A}\big[ F_{\scal}(\bw_k^{(\tau)})\big]
    \leq L\frac{1}{m}\sum_{k=1}^{m}\|\bw_{k}^{(\tau)}-{\overline{\bw}^{(\tau)}}\|_{2}^{2}
    + \frac{\eta L^{2}}{m}
    \sum_{\nu=0}^{\tau}\sum_{k=1}^{m}\|{\overline{\bw}^{(\nu)}}-\bw_{k}^{(\nu)}\|_{2}^{2}
    +\frac{L}{m} \eta^{2} \tau B^2.
\end{equation}
This inequality would suffice to prove that the distributed on-average stability increase with the accumulation of the “consensus distance”.

Suppose that the consensus distance is controlled below the critical consensus distance $\Gamma^2$ from $t_\Gamma$-th iterate to the end of the training. For simplicity, we make a mild assumption that the consensus distance $ \Gamma^2\leq\frac{1}{m}\sum_{k=1}^{m}\|\bw_{k}^{(t)}-{\overline{\bw}^{(t)}}\|_{2}^{2}\leq K^2$ if $t\leq t_\Gamma$. Therefore, the averged empirical risk at $\tau$-th iterate can be bounded as
\begin{align}
    \frac{1}{m}\sum_{k=1}^{m}\ebb_{\scal,A}\big[ F_{\scal}(\bw_k^{(\tau)})\big]
    &\leq L\Gamma^2
    + \frac{\eta L^{2}}{m}
    \sum_{\nu=0}^{t_\Gamma}\sum_{k=1}^{m}\|{\overline{\bw}^{(\nu)}}-\bw_{k}^{(\nu)}\|_{2}^{2}+ \eta L^{2}(\tau-t_\Gamma)\Gamma^2
    + \frac{L}{m} \eta^{2} \tau B^2\nonumber\\
    &\leq L\Gamma^2
    + \eta L^{2}\sum_{\nu=0}^{t_\Gamma}\big(\frac{1}{m}\sum_{k=1}^{m}\|{\overline{\bw}^{(\nu)}}-\bw_{k}^{(\nu)}\|_{2}^{2}-\Gamma^2\big)+ (\eta L^{2}\Gamma^2
    + \frac{L}{m} \eta^{2}  B^2)\cdot \boldsymbol{\tau}\nonumber\\
    &\leq L\Gamma^2
    + \eta L^{2}(t_\Gamma+1)\big(K^2-\Gamma^2\big)+ (\eta L^{2}\Gamma^2
    + \frac{L}{m} \eta^{2}  B^2)\cdot \boldsymbol{\tau},
\end{align}
if $\tau$ is greater than $t_\Gamma$; and 
\begin{align}
    \frac{1}{m}\sum_{k=1}^{m}\ebb_{\scal,A}\big[ F_{\scal}(\bw_k^{(\tau)})\big]
    &\leq L\frac{1}{m}\sum_{k=1}^{m}\|\bw_{k}^{(\tau)}-{\overline{\bw}^{(\tau)}}\|_{2}^{2}
    + \frac{\eta L^{2}}{m}
    \sum_{\nu=0}^{\tau}\sum_{k=1}^{m}\|{\overline{\bw}^{(\nu)}}-\bw_{k}^{(\nu)}\|_{2}^{2}
    + \frac{L}{m} \eta^{2}  B^2\tau\nonumber\\
    & \leq L K^2
    + (\eta L^2K^2 +\frac{L}{m} \eta^{2}  B^2)\cdot \boldsymbol{\tau},
\end{align}
if if $\tau$ is smaller than $t_\Gamma$.

Consequently, 
\begin{align}
    G(t_\Gamma)=\sum_{\tau=0}^{t} &{C}^{t-\tau}\frac{1}{m}\sum_{k=1}^{m}\ebb_{\scal,A}\big[ F_{\scal}(\bw_k^{(\tau)})\big]
    = (\sum_{\tau=0}^{t_\Gamma}+\sum_{\tau=t_\Gamma+1}^{t}) {C}^{t-\tau}\frac{1}{m}\sum_{k=1}^{m}\ebb_{\scal,A}\big[ F_{\scal}(\bw_k^{(\tau)})\big]\nonumber\\
    \leq &\frac{L}{m} \eta^{2}  B^2\sum_{\tau=0}^{t}\tau {C}^{t-\tau} 
    + \eta L^2 K^2 \sum_{\tau=0}^{t_\Gamma} \tau {C}^{t-\tau}
    + \eta L^2 \Gamma^2 \sum_{\tau=t_\Gamma+1}^{t} \tau {C}^{t-\tau}\nonumber\\
     & +  LK^2(\sum_{\tau=0}^{t_\Gamma}{C}^{t-\tau})
    + [L\Gamma^2
    + \eta L^{2}(t_\Gamma+1)\big(K^2-\Gamma^2\big)](\sum_{\tau=t_\Gamma+1}^{t}{C}^{t-\tau}).
\end{align}
Recall that \textbf{our goal} is to prove that $G(t_\Gamma)$ increase with $t_\Gamma$.

Due to the fact that $\sum_{\tau=0}^{t}\tau {C}^{t-\tau}=\sum_{\tau=0}^{t}(t-\tau) {C}^{\tau}\leq t\sum_{\tau=0}^{t}{C}^{\tau}$, we can obtain
\begin{align}
    G(t_\Gamma) \leq & \frac{t\frac{L}{m} \eta^{2}  B^2}{1- C} 
    + \eta L^2 K^2 \sum_{\tau=0}^{t_\Gamma} \tau {C}^{t-\tau}
    + \eta L^2 \Gamma^2 \sum_{\tau=t_\Gamma+1}^{t} \tau {C}^{t-\tau}\nonumber\\
    & +  \frac{C^t LK^2}{1- C}
    + [L\Gamma^2
    + \eta L^{2}(t_\Gamma+1)\big(K^2-\Gamma^2\big)]\frac{C^t C^{t_\Gamma+1}}{1- C}.
\end{align}
Since the finite sum of the arithmetico-geometric sequence can be written as
\begin{equation}
    \sum_{\tau=0}^{t}\tau{C}^{-\tau} = [\frac{t}{C^{-1}-1}-\frac{1}{(C^{-1}-1)^2}]C^{-(t+1)}+[\frac{1}{(C^{-1}-1)^2}+\frac{1}{C^{-1}-1}],
\end{equation}
we can upper bound $G(t_\Gamma)$ as follows:
\begin{align}
    G(t_\Gamma) \leq & \frac{t\frac{L}{m} \eta^{2}  B^2}{1- C} 
    + \eta L^2 \Gamma^2 C^t \{[\frac{t}{C^{-1}-1}-\frac{1}{(C^{-1}-1)^2}]C^{-(t+1)}+[\frac{1}{(C^{-1}-1)^2}+\frac{1}{C^{-1}-1}]\} \nonumber\\
    & + \eta L^2 (K^2-\Gamma^2) C^t \{[\frac{t}{C^{-1}-1}-\frac{1}{(C^{-1}-1)^2}]C^{-(\boldsymbol{t_\Gamma}+1)}+[\frac{1}{(C^{-1}-1)^2}+\frac{1}{C^{-1}-1}]\} \nonumber\\
    & +  \frac{C^t LK^2}{1- C}
    + [L\Gamma^2
    + \eta L^{2}(\boldsymbol{t_\Gamma}+1)\big(K^2-\Gamma^2\big)]\frac{C^t C^{\boldsymbol{t_\Gamma}+1}}{1- C}.
\end{align}
Rewrite the inequality above, then we arrive at
\begin{align}
    G(t_\Gamma) \leq & \frac{t\frac{L}{m} \eta^{2}  B^2}{1- C} 
    \unaryplus \eta L^2 \Gamma^2 C^t [\frac{t}{C^{-1}-1}\unaryminus\frac{1}{(C^{-1}-1)^2}]C^{-(t+1)}
    \unaryplus  \frac{C^t LK^2}{1- C} \nonumber\\
    & + \eta L^2 (K^2-\Gamma^2) C^t\{ [\frac{t}{C^{-1}-1}\unaryminus\frac{1}{(C^{-1}-1)^2}] C^{-(\boldsymbol{t_\Gamma}}+1) +  \boldsymbol{t_\Gamma}\frac{C^{\boldsymbol{t_\Gamma}+1}}{1- C} \}
    + [L\Gamma^2
    + \eta L^{2}\big(K^2-\Gamma^2\big)]C^t\frac{C^{\boldsymbol{t_\Gamma}+1}}{1- C}.
\end{align}
One can prove that if $t\geq \frac{-C}{2\ln{C}}$, the upper bound of $G(t_\Gamma)$ will be a monotonically increasing function of $t_\Gamma$.
Consequently, we can conclude that the distributed on-average stability bound and the generalization bound of D-SGD increase monotonically with $t_\Gamma$ if the total number of iterations satisfies $t\geq \frac{-C}{2\ln{C}}$.

{\color{magenta}\qed}